\documentclass{article}

\usepackage[preprint]{neurips_2026}

\usepackage[utf8]{inputenc} %
\usepackage[T1]{fontenc}    %
\usepackage{hyperref}       %
\usepackage{url}            %
\usepackage{booktabs}       %
\usepackage{amsfonts}       %
\usepackage{nicefrac}       %
\usepackage{microtype}      %
\usepackage{xcolor}         %
\usepackage{amsmath}
\usepackage{makecell}
\usepackage{wrapfig}
\usepackage{cleveref}
\usepackage{graphicx}
\usepackage{multirow}
\usepackage{subcaption}
\usepackage{authblk}
\usepackage{float}
\usepackage{tabularx}
\usepackage{array}
\usepackage{siunitx}
\title{TokenMatch: 3D Mesh Correspondence Transformer with Curvature-Guided Tokenisation} 

\author{
\textbf{Adeela Islam}\textsuperscript{1,2,6,}\thanks{work done during internship at the 4DQV research group, Max Planck Institute for Informatics}\quad
\textbf{Zorah L\"ahner}\textsuperscript{3,4}\quad
\textbf{Vittorio Murino}\textsuperscript{1,5}\quad
\textbf{Vladislav Golyanik}\textsuperscript{6,}\thanks{corresponding author’s email: \texttt{golyanik@mpi-inf.mpg.de}}\\

\textsuperscript{1}Italian Institute of Technology \quad
\textsuperscript{2}University of Genoa\quad
\textsuperscript{3}University of Bonn\\
\textsuperscript{4}Lamarr Institute\quad
\textsuperscript{5}University of Verona\quad
\textsuperscript{6}Max Planck Institute for Informatics, SIC
\\\vspace{10pt}
\definecolor{hblue}{HTML}{2F6DA1}
\textcolor{hblue}{\url{https://4dqv.mpi-inf.mpg.de/TokenMatch/}}
}

\begin{document}

\maketitle

\begin{abstract}
  While data-driven 3D shape correspondence estimation has recently seen substantial progress, robust matching under partial observations and strong non-isometric deformations remains challenging. Existing learning-based approaches often rely on hand-crafted descriptors or template-based representations, whereas recent generative models over functional maps suffer from high inference cost, limited interpretability, and poor generalisation to partial shapes. 
  In response to these limitations, this paper introduces TokenMatch, a new transformer-based unified model for estimating 3D shape correspondences. 
  Our feed-forward approach trained exclusively on BeCoS, a challenging non-isometric partial-to-partial shape-matching dataset, can generalise to matching full shapes without retraining or fine-tuning. 
  TokenMatch uses self- and cross-attention mechanisms to efficiently learn patch-level and point-level relations as well as dense correspondences between shape pairs. 
  Our core insight is that meshes can be adaptively tokenised into patches using shape curvature guidance, enabling effective learning of shape-specific geometric descriptors for correspondence estimation. 
  We evaluate TokenMatch on standard benchmarks for partial and full shape matching, including CP2P, PSMAL, BeCoS, FAUST, SCAPE, and SHREC'19. 
  Our method achieves consistently high performance, in most cases outperforming existing methods for partial and full shape matching in the mean geodesic error and intersection-over-union metrics, while also running faster at sub-second inference speeds. 
\end{abstract}

\begin{figure}[!t]
  \centering
  \includegraphics[width=\linewidth]{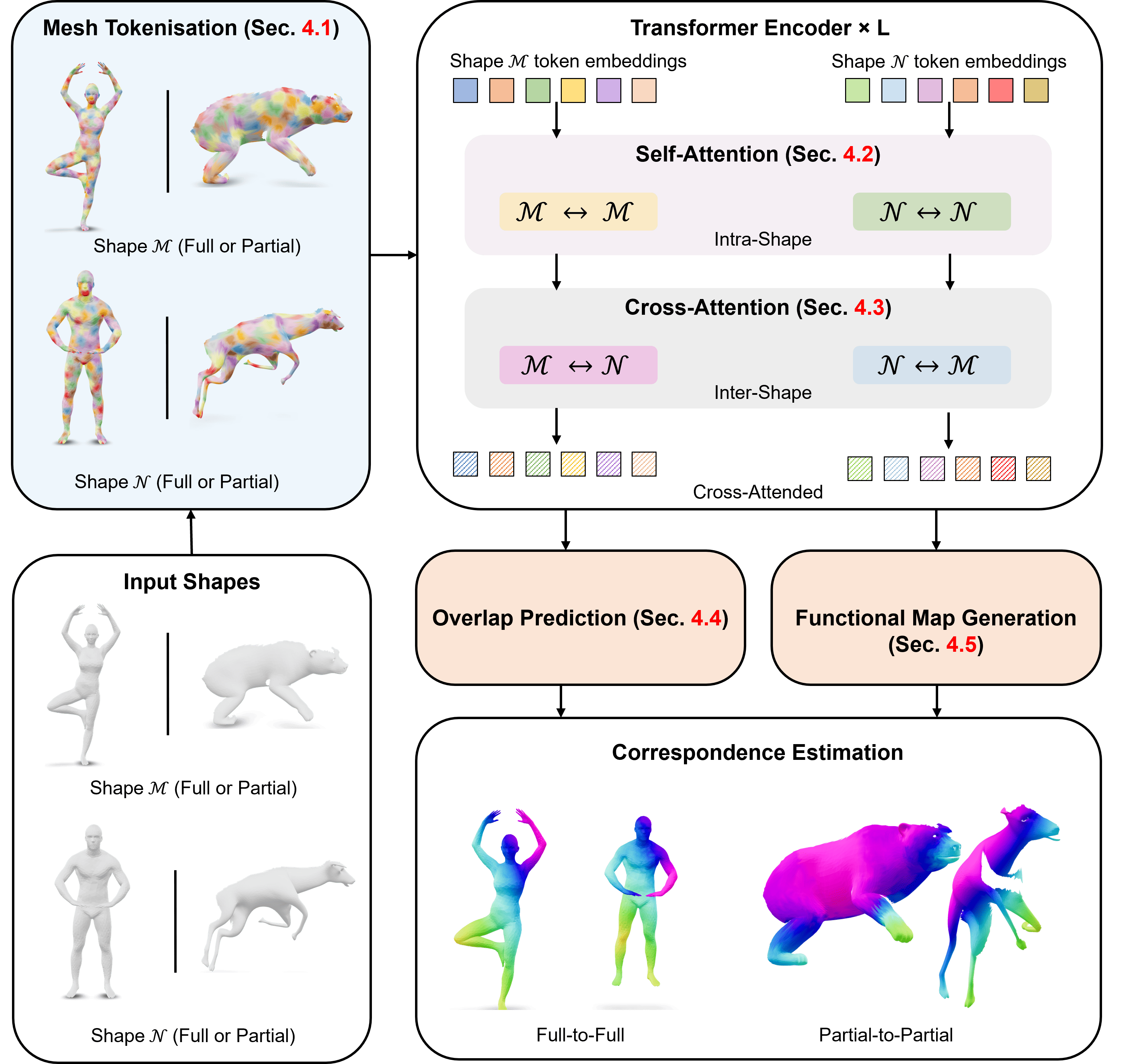}
  \vspace{-2pt}
  \caption{\textbf{Overview of TokenMatch, our unified model for 3D mesh correspondence estimation.} Given a pair of input meshes (bottom-left), we extract geometry-aware tokens (top-left; Sec.~\ref{sec:token}) and process them with a transformer encoder to obtain shape features (top-right; Secs.~\ref{sec:mae} and \ref{sec:overlap}). 
  Cross-attention enables efficient learning of inter-shape interactions, and the resulting representations are used to estimate functional maps. 
  An additional overlap prediction module identifies shared regions between shapes (Sec.~\ref{sec:overlap}). 
  Our method is first pre-trained in a self-supervised manner using a masked reconstruction objective and then trained on annotated partial-shape matching data using PointInfoNCE loss (Sec.~\ref{sec:cross-attn}) and 
  functional map supervision (Sec.~\ref{sec:fm}). 
  }
  \label{fig:overview}
  \vspace{-10pt}
\end{figure}

\section{Introduction} \label{sec:intro}

Establishing accurate correspondences between 3D shapes is a fundamental and open problem in geometry processing~\cite{bronstein2010scale,cao2023robust,eisenberger2020smooth,marin2024nicp,ovsjanikov2012functional,benkner2021q,sun2023spatially}, with applications including texture transfer~\cite{dinh2005texture, ShimadaDispVoxNets2019}, segmentation~\cite{rustamov2007laplace}, deformation transfer~\cite{sumner2004deformation}, and statistical shape modelling~\cite{Allen2003, Hasler2009}. 
Given two shapes, the goal of shape correspondence estimation is to identify semantically consistent pointwise correspondences despite variability in pose, sampling, noise, and, critically, degree of partiality~\cite{zhuravlev2026nonrigid}. 
Among existing approaches, the functional map framework~\cite{ovsjanikov2012functional,xie2026deepshapematchingkit} has emerged as a powerful paradigm, representing correspondences as compact linear operators in the Laplace--Beltrami spectral domain. 
Despite substantial progress in full-to-full shape matching~\cite{zhuravlev2025denoising, eisenberger2020smooth, ezuz2019elastic}, partial-shape matching remains a major open challenge: shapes may only partially overlap, increasing the solution space, making the problem particularly ill-posed, and often also requiring identification of the shared region.
In addition, existing learning-based approaches often rely on assumptions of complete shapes or near-isometric deformations~\cite{attaiki2021dpfm,ehm2024geometrically,xie2025echomatch,pierson2025diffumatch}, and are typically trained on small-scale and single-category datasets. As a result, they can encode dataset-specific biases that limit robustness to unseen deformations and artefacts and generalisation across partial and full shape matching settings~\cite{ehm2026integer}.
We address several open challenges in the field by introducing a new unified model that serves as a foundation for shape correspondence estimation across different settings; see Fig.~\ref{fig:overview}. 
Our method, which we call \textit{TokenMatch}, is based on a transformer encoder. 
Transformers have recently emerged as a dominant paradigm for representation learning due to their ability to model long-range dependencies and flexible interactions between input elements~\cite{vaswani2017attention,dosovitskiy2021an,yu2025pixeldit}.
We argue that their ability to model relationships between variable-sized sets of elements makes them particularly well-suited for partial-shape matching, where the number and arrangement of input regions are not known in advance. 
At the same time, while significant progress has been made in adapting transformers to 3D data, applying them to shape correspondence estimation is non-trivial due to the irregular structure of meshes and the absence of a natural and universal tokenisation. 
Hence, the core ingredient of our mesh transformer for 3D shape correspondence is a new \textit{curvature-guided tokenisation scheme}. 
Our idea is to represent shapes as collections of geometry-aware surface patches, which serve as transformer tokens. 
The construction of these patches is guided 
using local curvature and spectral shape cues, capturing both local geometric detail and intrinsic shape structure. 
Within our formulation, self-attention models intra-shape geometric relationships and interactions, while cross-attention explicitly captures dense correspondence between shapes. 
As a result, the designed mesh transformer architecture jointly reasons about geometry and correspondence through attention mechanisms. 
We choose functional maps \cite{ovsjanikov2012functional}, i.e., spectral representation for encoding and recovering pointwise correspondences from learned features. 
Importantly, TokenMatch is template-free and does not rely on category-specific assumptions, making it applicable across diverse shape categories. 
Since our method accepts a varying number of tokens per shape, it generalises naturally to full shapes, while not necessarily requiring training or fine-tuning on them (though it can be done to improve performance). 
We train TokenMatch on the newly introduced large-scale,  curated BeCoS dataset with multiple shape categories \cite{ehm2025beyond}, which we consider a key factor behind its performance, alongside curvature-guided tokenisation. 
Notably, BeCoS provides diverse full and partial shape categories (humanoids and animals) with challenging, highly non-isometric deformations at realistic scales, encouraging robustness to deformation patterns that go beyond standard nearly isometric settings. 
It enables method training, evaluation and---as we demonstrate experimentally in Sec.~\ref{sec:experiments}---cross-category zero-shot generalisation of TokenMatch across more than $2500$ unique shapes. 
Last but not least, TokenMatch is a feed-forward approach that achieves sub-second inference time, which is unprecedented in partial-shape matching, especially for combinatorial techniques and when no additional assumptions (e.g., on mesh discretisation) are made. 
To summarise, the \textbf{main technical contributions} of this paper are as follows: 
\vspace{-3pt}
\begin{itemize}
    \item TokenMatch, the first mesh-transformer-based \emph{unified model} for 3D shape correspondence estimation, combining functional maps with self- and cross-attention mechanisms to model geometric relationships and dense correspondences between shape pairs (Sec.~\ref{sec:method}). 
    Our template-free formulation does not rely on shared templates or category-specific assumptions, enabling broader applicability across shape domains. 
    \item A curvature-guided, geometry-aware, overlapping mesh tokenisation that enables transformer-based processing of irregularly sampled shapes, i.e., meshes with spatially varying sampling density (Sec.~\ref{sec:token}). 
\end{itemize}
To further improve robustness and generalisation of TokenMatch, we employ a masked auto-encoder pre-training strategy, encouraging the model to learn geometry-aware features that are resilient to noise, partiality, and varying sampling densities (Sec.~\ref{sec:mae}). 
To enable correspondence-aware reasoning, we employ cross-attention between features of shape pairs (Sec.~\ref{sec:cross-attn}), and we also predict overlaps between the shapes (Sec.~\ref{sec:overlap}). 
In the general case, our TokenMatch approach outputs partial functional maps (Secs.~\ref{sec:fm} and \ref{ssec:final_training}). 
We plan to release the source code and trained models for research purposes to facilitate reproducibility. 
\section{Related Work} \label{sec:rw}
This section reviews methods closely related to ours. 
For a broader overview of the field, we refer to the surveys on shape correspondence estimation and registration~\cite{zhuravlev2026nonrigid, deng2022survey, van2011survey}.   

\subsection{Shape Correspondence Estimation and 3D Transformer Models} 
Spectral methods for correspondence estimation \textbf{between full meshes} learn pointwise descriptors that induce functional maps~\cite{ovsjanikov2012functional,ovsjanikov2016computing}, yielding a compact low-rank representation in the Laplace--Beltrami eigenbasis. 
Deep learning has further enabled learning descriptor functions directly from data~\cite{litany2017deep,sharp2022diffusionnet}, evolving from supervised formulations~\cite{donati2020deep,sharma2020weakly} to unsupervised methods~\cite{cao2023robust,roufosse2019unsupervised}. 
This has led to a range of functional-map-based pipelines~\cite{zhuravlev2025denoising,bastian2024hybrid,halimi2019unsupervised}, often augmented with spatial information to improve robustness~\cite{eisenberger2020smooth,sundararaman2024deformation,bastian2024hybrid}. 
However, the global low-rank structure of functional maps limits fine-grained correspondence recovery, and many methods therefore rely on independently computed per-shape features followed by a separate alignment stage~\cite{rodolavmv15}. 
\textbf{Partial-shape matching}  
represents one of the central challenges in the field. 
This variant considers 
missing geometry and reduced global structure, requiring both pointwise correspondence estimation and identification of overlapping regions. 
Combinatorial approaches to partial-shape matching are effective under controlled settings, but do not scale well to high-resolution shapes due to their combinatorial computational complexity \cite{roetzer2022scalable}. 
Unsupervised methods remain fragile due to limited regularisation and are particularly sensitive to topological noise such as self-intersections or reconstruction artefacts~\cite{rodola2017partial, ehm2025beyond, mei2023overlap}.
To improve efficiency, functional map-based methods extend spectral correspondence to partial settings. 
DPFM~\cite{attaiki2021dpfm} introduces an overlap prediction module (vertex-wise) and 
EchoMatch~\cite{xie2025echomatch} enforces cycle consistency of pointwise maps as a form of correspondence verification, while also demonstrating the effectiveness of pre-trained features such as DINOv2~\cite{oquab2023dinov2} for partial matching.
These methods often struggle to reliably infer overlapping regions.
Their reliance on global spectral structure, vertex-level interactions or explicit overlap heuristics limits their ability to jointly reason about geometry and correspondence at scale. 
Transformers~\cite{vaswani2017attention} have emerged as a dominant architecture in modern deep learning due to their ability to act as a highly scalable, differentiable, general-purpose computational paradigm with minimal inductive bias. By relying on token interactions through self-attention, they naturally model long-range dependencies and global structure without imposing strong assumptions about the underlying signal. This property is particularly appealing for 3D learning, where geometric structure can vary significantly across resolution, topology, and sampling density. Consequently, transformer-based architectures have recently shown strong performance in a range of 3D tasks, including point cloud understanding~\cite{hu2022subdivision} and mesh-based representation learning~\cite{meshmae2022}.

More recent advances further demonstrate the effectiveness of carefully designed transformer backbones for 3D data. In particular, Point Transformer V3~\cite{wu2024point} has shown that incorporating geometry-aware attention mechanisms while preserving the general-purpose nature of transformers leads to strong performance and scalability on large-scale point cloud tasks. However, many existing approaches for shape correspondence either rely on vertex-level tokenisation or operate in learned latent shape spaces~\cite{trappolini2021shape}, which can obscure fine-grained geometric structure and limit robustness under partial or non-uniform sampling. 
In contrast, we propose a curvature-driven mesh tokenisation that preserves intrinsic geometric cues while maintaining compatibility with a transformer backbone.
The functional map framework combined with transformer-based modelling efficiently captures both spectral shape structure and cross-shape interactions. 
This enables the first mesh transformer for correspondence estimation that unifies full and partial 3D shape matching within a single framework, leveraging the flexibility and global reasoning capability of transformers while better respecting local surface geometry.
Compared to partial-shape matching methods \cite{ehm2024partial, roetzer2022scalable, attaiki2021dpfm, xie2025echomatch}, our feed-forward transformer-based formulation achieves the fastest inference among the evaluated methods, with substantial speed-up over the optimisation-based approaches \cite{ehm2024partial, roetzer2022scalable}. 
\subsection{Data-Driven Methods with Generative Priors} 
Denoising diffusion probabilistic models (DDPMs) have emerged as a leading paradigm for generative modelling, formulating 2D image synthesis as an iterative denoising process that inverts a gradual corruption with Gaussian noise, often in a conditional setting~\cite{sohl2015deep, ho2020denoising, po2024state}. 
The 2D structure of functional maps~\cite{ovsjanikov2012functional}
has motivated connections to generative models, enabling DDPMs to operate on low-dimensional correspondence representations~\cite{cohen2025fridu}. 
Recent works explore this direction in different ways and show alternatives to explicit functional map solvers. 
DenoisFM~\cite{zhuravlev2025denoising} directly generates functional maps and mitigates Laplacian eigenfunction sign ambiguity via an unsupervised selection strategy based on learned surface features, but relies on pre-defined category-specific shape templates. 
DiffuMatch~\cite{pierson2025diffumatch} operates on absolute-valued functional maps to avoid sign ambiguity, replacing classical constraints such as orthogonality and Laplacian commutativity with learned regularisation. 
Despite their effectiveness, these approaches introduce significant computational overhead due to iterative sampling and rely on large-scale data to learn reliable priors over the functional map space. 
Their dependence on category-specific training distributions may limit diversity of the generated samples and cross-category generalisation.
Hence, our approach is distantly related to DenoisFM and DiffuMatch in that it is data-driven; in contrast to them, we leverage an efficient feed-forward transformer model and its inductive biases for estimating shape correspondences. 
Moreover, compared to DenoisFM, our approach is template-free and is orders of magnitude faster than slow guided DDPM generation. 

\newpage
\section{Definitions and Background}\label{sec:background}

This section provides core definitions used in this paper and reviews functional maps. 

\textbf{Definitions.} Let $\mathcal{M}$ and $\mathcal{N}$ be two 3D shapes represented as triangular meshes, with vertex sets $V_{\mathcal{M}}$ and $V_{\mathcal{N}}$ of sizes $v_{\mathcal{M}} = |V_{\mathcal{M}}|$ and $v_{\mathcal{N}} = |V_{\mathcal{N}}|$, respectively. Our goal is to estimate a pointwise mapping
$\Pi_{\mathcal{N}\mathcal{M}} : \mathcal{M} \rightarrow \mathcal{N}$,
which assigns each point on $\mathcal{M}$ to a corresponding point on $\mathcal{N}$; in the partial setting, the mapping is defined only on a subset of $\mathcal{M}$.
Direct estimation of $\Pi_{\mathcal{N}\mathcal{M}} \in \{0,1\}^{v_{\mathcal{N}} \times v_{\mathcal{M}}}$ is computationally expensive. Therefore, we adopt the functional map framework \cite{ovsjanikov2012functional}, where correspondences are represented in a reduced spectral basis. 

\textbf{Functional maps.} 
For each shape $\mathcal{S} \in \{\mathcal{M}, \mathcal{N}\}$, we compute the Laplace--Beltrami operator $\mathbf{L}_{\mathcal{S}}$ and its first $k$ eigenvectors 
$\boldsymbol{\Phi}_{\mathcal{S}} \in \mathbb{R}^{v_{\mathcal{S}} \times k}$ so that 
$\mathbf{L}_{\mathcal{S}} \boldsymbol{\Phi}_{\mathcal{S}} = \boldsymbol{\Phi}_{\mathcal{S}} \boldsymbol{\Lambda}_{\mathcal{S}}$,
where $\boldsymbol{\Lambda}_{\mathcal{S}}$ contains the eigenvalues. The functional map $\mathbf{C}_{\mathcal{M}\mathcal{N}} \in \mathbb{R}^{k \times k}$ is defined as:
\begin{equation}
\mathbf{C}_{\mathcal{M}\mathcal{N}} = \boldsymbol{\Phi}_{\mathcal{N}}^{\dagger} \, \Pi_{\mathcal{N}\mathcal{M}} \, \boldsymbol{\Phi}_{\mathcal{M}}, 
\end{equation}
where ``$(\cdot)^{\dagger}$'' denotes Moore--Penrose pseudoinverse. 
The learned feature descriptors $\mathbf{F}_{\mathcal{M}}$ and $\mathbf{F}_{\mathcal{N}}$ are projected onto the spectral domain, resulting in $\mathbf{A}_{\mathcal{M}}$ and $\mathbf{A}_{\mathcal{N}}$: 
\begin{equation}
\mathbf{A}_{\mathcal{M}} = \boldsymbol{\Phi}_{\mathcal{M}}^{\dagger} \mathbf{F}_{\mathcal{M}}, 
\quad\text{and}\quad
\mathbf{A}_{\mathcal{N}} = \boldsymbol{\Phi}_{\mathcal{N}}^{\dagger} \mathbf{F}_{\mathcal{N}}.
\end{equation}
The functional map~\cite{ovsjanikov2012functional} is obtained as a solution to the following optimisation problem:
\begin{equation}
\mathbf{C}_{\mathcal{M}\mathcal{N}} = \arg\min_{\mathbf{C}} 
\left\| \mathbf{C} \mathbf{A}_{\mathcal{M}} - \mathbf{A}_{\mathcal{N}} \right\|_F^2
+ \lambda \left\| \mathbf{C} \boldsymbol{\Lambda}_{\mathcal{M}} - \boldsymbol{\Lambda}_{\mathcal{N}} \mathbf{C} \right\|_F^2,
\end{equation}
with a scalar weight $\lambda$ and Frobenius norm denoted by ``$\left\|\cdot\right\|_F$''. 
The second term enforces commutativity with the Laplace--Beltrami operators~\cite{ren2019structured}. Our goal is to learn robust feature descriptors that lead to accurate functional maps. 

\section{TokenMatch: Our 3D Mesh Correspondence Transformer}\label{sec:method}

We propose a transformer-based, unified framework as a foundation for partial and full 3D shape correspondence estimation. 
Given two triangular meshes $\mathcal{M} = (V_{\mathcal{M}}, T_{\mathcal{M}})$ and $\mathcal{N} = (V_{\mathcal{N}}, T_{\mathcal{N}})$, our goal is to predict accurate point-wise correspondences between their vertices $V_{\mathcal{M}}, V_{\mathcal{N}}$ in a feed-forward manner. 
Unlike approaches that rely on fixed spectral representations or handcrafted descriptors~\cite{ovsjanikov2012functional,ren2019structured,donati2020deep}, we operate directly on tokenised mesh geometry. 
This design avoids committing to a fixed representation or matching pipeline, while enabling scalable joint modelling of local surface geometry and global shape context.
It also allows us to leverage a large-scale dataset. 
\begin{wrapfigure}{r}{0.5\textwidth}
  \centering
  \vspace{-15pt} %
  \includegraphics[width=\linewidth]{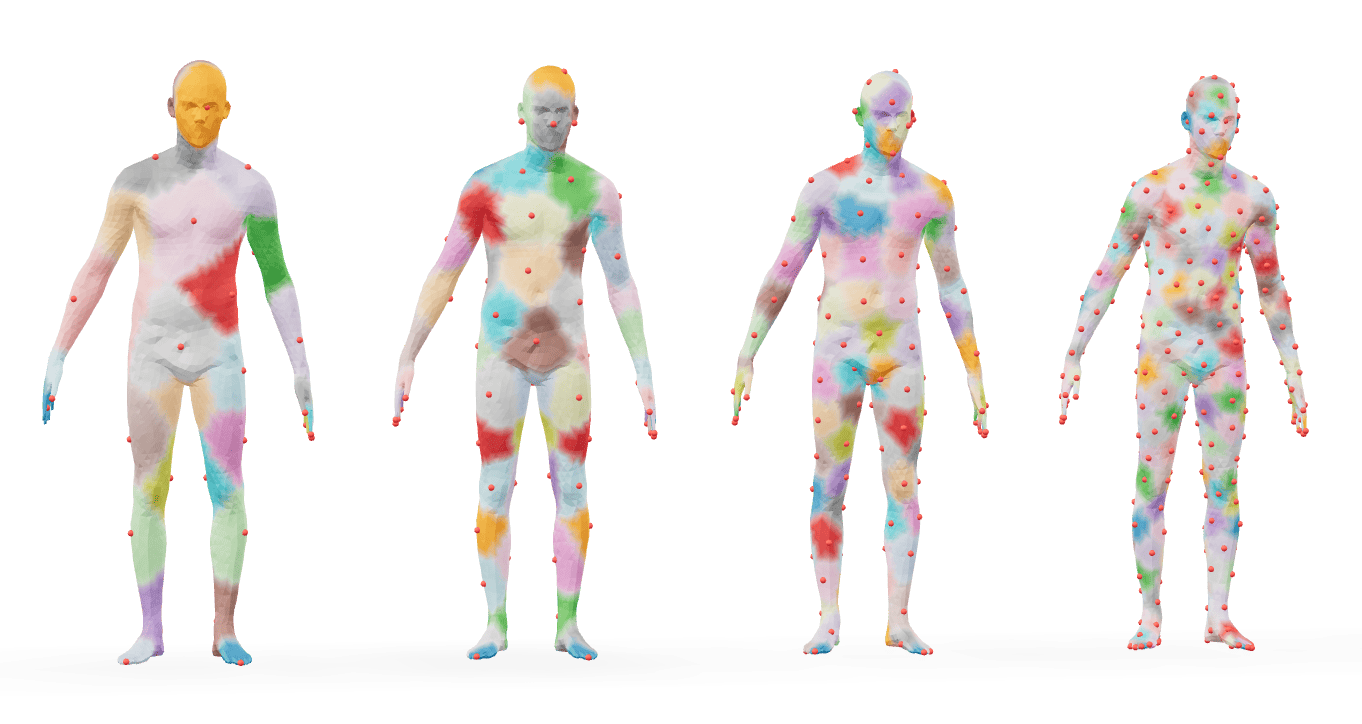}
  \vspace{-20pt}
  \caption{\textbf{Effect of increasing token density.} The number of tokens increases from left to right (32, 64, 128, 256). Red dots indicate the token centres. Increasing the number of tokens improves coverage of critical regions, first capturing fingers and later extending to toes. Zoom recommended.} 
  \label{fig:mesh-tokenization}
 \vspace{-10pt} 
\end{wrapfigure}
To this end, we adapt the mesh representation into a transformer-friendly format inspired by masked mesh and point cloud auto-encoders \cite{meshmae2022, Li2023}, enabling effective patch-wise processing of surface geometry. Our method learns expressive features to estimate functional maps, providing a robust solution for non-rigid and non-isometric shape correspondence. Fig.~\ref{fig:overview} provides an overview of our framework. 
We first perform geometry-aware tokenisation of the input mesh (Sec.~\ref{sec:token}). 
These tokens are then processed by a transformer encoder that learns contextualised shape features through pre-training (Sec.~\ref{sec:mae}). 
As independent per-shape features lack cross-shape awareness, we introduce cross-attention for feature exchange between shapes (Sec.~\ref{sec:cross-attn}), use the resulting features to predict overlap regions (Sec.~\ref{sec:overlap}) and, finally, estimate functional maps (Secs.~\ref{sec:fm} and \ref{ssec:final_training}). 

\begin{figure}[t]
    \centering
    \includegraphics[width=0.9\linewidth]{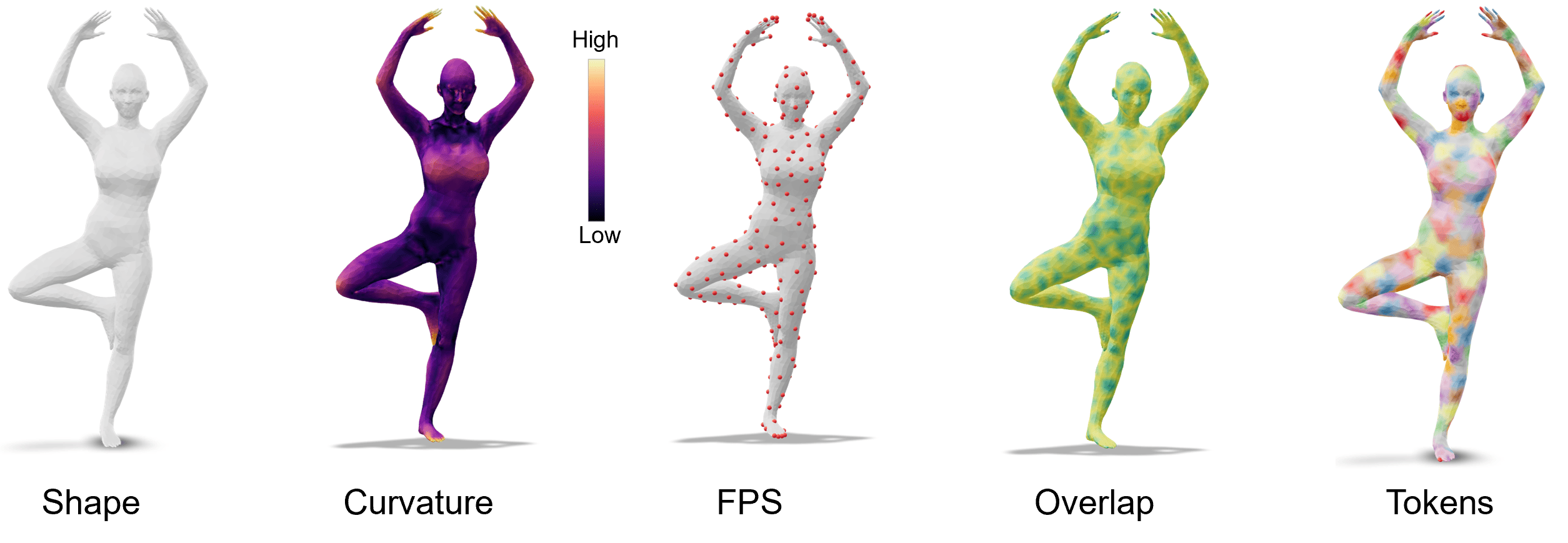}
    \caption{\textbf{The proposed geometry-aware mesh tokenisation:}
    The process begins with a triangular input mesh representing a human pose. A spectral signal is computed and visualised as a heatmap. Points are sampled using curvature-weighted farthest point sampling (FPS). Soft overlapping regions are then defined using a geodesic influence radius, and mesh vertices are softly assigned to token regions. Final tokens are represented by their sampled centres.} 
    \vspace{-5pt}
    \label{fig:mesh-tokenization-pipeline}
\end{figure}

\subsection{Geometry-Aware Curvature-Guided Mesh Tokenisation}\label{sec:token}

Given a triangular mesh $\mathcal{S} = (V_{\mathcal{S}}, T_{\mathcal{S}})$, where $V_{\mathcal{S}}$ denotes the set of vertices and $T_{\mathcal{S}}$ denotes mesh connectivity, we construct $g$ geometry-aware tokens for transformer processing. We partition the mesh into overlapping regions to capture smoothly varying geometric structure. 
As discussed in App.~\ref{subsec:overlap-degree}, moderate token overlap is important for accurate shape correspondence (see also Table~\ref{tab:sigma_ablation}). 
To guide tokenisation, we compute a per-vertex signal combining local and spectral geometry:
\begin{equation}
s(i) = \alpha \left\lVert \mathbf{H}(i) \right\rVert_2
       + (1-\alpha)E(i),
\end{equation}
where $i \in V_{\mathcal{S}}$, $\mathbf{H}(i)$ denotes the absolute mean curvature at vertex $i$, and is therefore non-negative. $\alpha \in [0,1]$ balances local geometric detail and global spectral structure, and $E(i)$ denotes the spectral energy derived from the Laplace--Beltrami eigenfunctions. 
We set \(\alpha\) to $0.6$. 
Next, let $\phi_{\ell}(i)$ denote the value of the $\ell$-th eigenfunction at vertex $i$. We compute the spectral energy as $E(i) = \sum_{\ell=5}^{16} \phi_{\ell}(i)^2$. This mid-frequency band captures stable structural details while excluding low-frequency modes ($\ell < 5$), which mainly encode global shape information, and high-frequency modes ($\ell > 16$), which are more sensitive to remeshing and geometric noise. 
\textbf{Curvature-aware sampling.} 
We select $g$ token centres using farthest point sampling (FPS) \cite{Eldar1997} with a signal-weighted geodesic distance. Let $\mathcal{C}_t$ denote the set of selected centres after iteration $t$, with $\mathcal{C}_0 = \emptyset$. At iteration $t$, the next centre is chosen as follows: 
\begin{equation}
c_t = \arg\max_{i \in V_{\mathcal{S}}}
\left(
\min_{c \in \mathcal{C}_{t-1}} d_{\mathrm{geo}}(i,c)\cdot (1 + \beta s(i))
\right),
\end{equation}
where $d_{\mathrm{geo}(i, c)}$ denotes geodesic distance between vertices $i$ and $c$, $s(i)$ is the geometry-aware vertex  signal, and $\beta > 0$ controls the influence of the geometry-aware signal.  
We choose \(\beta=0.5\). 

\textbf{Soft token assignment.} 
Given sampled centres $\{c_i\}_{i=1}^{g}$ and mesh element $x$,  
we define a soft assignment of mesh elements to tokens as follows:
\begin{equation}
\label{eq:soft_assignment}
w_i(x) =
\frac{\exp\left(-d_{\mathrm{geo}}(c_i,x)^2 / 2\sigma^2\right)}
{\sum_{k=1}^{g} \exp\left(-d_{\mathrm{geo}}(c_k,x)^2 / 2\sigma^2\right)}, 
\end{equation}
where \( w_i(x) \) denotes the soft assignment weight of mesh element \( x \) to token \( i \), representing the degree to which \( x \) belongs to the token region centred at \( c_i \). 
This formulation induces overlapping token regions governed by the geodesic Gaussian bandwidth and $g$. 
A broader bandwidth $\sigma$ produces more diffuse tokens with stronger overlap, while increasing $g$ reduces spacing between token centres, further enhancing overlap. 
According to Eq.~\eqref{eq:soft_assignment}, each $x$ contributes to all tokens with \( w_i(x) \) dependent on $d_{\mathrm{geo}}$, and we do not impose any truncation distance threshold\footnote{for \(\sigma=0\), we use hard assignment, i.e., each mesh element is assigned to its nearest token centre in geodesic distance}. 
In an ablation study in App.~\ref{subsec:overlap-degree}, we analyse how these design choices affect the final performance (see Table~\ref{tab:sigma_ablation}). 
As shown in Figs.~\ref{fig:mesh-tokenization} and \ref{fig:mesh-tokenization-pipeline}, the proposed soft tokenisation offers several advantages over hard surface partitioning: Its overlapping design improves robustness under varying mesh resolution and sampling density, since token representations arise from weighted aggregation rather than rigid assignments. As a result, the representation becomes less sensitive to irregular tessellations and discretisation noise. Conceptually, this can be viewed as a geometry-aware weighting scheme analogous to attention over intrinsic distances, where influence decays smoothly with geodesic distance~\cite{bhardwaj2024curvature}. Such a formulation enables the transformer to operate on more expressive and geometry-consistent inputs, which is especially beneficial for partial and full-shape correspondence estimation. 
We further analyse our design choices in App.~\ref{subsec:token-design}, where we provide an ablation study on different tokenisation strategies. 
We demonstrate that our curvature-guided mesh tokenisation yields the most consistent performance across multiple benchmarks, validating its effectiveness compared to alternative mesh tokenisation approaches.

\subsection{Transformer Backbone}\label{sec:mae}
We now describe how the mesh tokens are processed by a transformer to obtain contextualised shape features.
Given the geometry-aware tokenisation, each mesh element $x \in \mathcal{S}$ is assigned to token regions via soft weights $w_i(x)$, as defined in Sec.~\ref{sec:token} and Eq.~\eqref{eq:soft_assignment}. The weights $w_i(x)$ are used to aggregate element-wise descriptors
into token features, producing geometric representations that capture both intrinsic and extrinsic surface cues. 
Each token is associated with its centre coordinate $\mathbf{c}_i \in \mathbb{R}^3$ obtained via curvature-aware FPS (Sec.~\ref{sec:token}) and encoded as $\mathbf{p}_i = \mathrm{MLP}(\mathbf{c}_i)$. The final token representation is given by $\mathbf{x}_i = \mathbf{e}_i + \mathbf{p}_i$, forming the transformer input sequence $\mathbf{H}^{(0)} = ({\mathbf{x}_1, \dots, \mathbf{x}_g})$. 
%
%
We use an MLP to project the aggregated feature vector into the token representation \(\mathbf{e}_i\). 
The token sequence is then processed through $L$ transformer layers
$\mathbf{H}^{(\ell+1)} = \mathrm{TransformerBlock}(\mathbf{H}^{(\ell)})$,
yielding the final contextualised token features $\mathbf{Z} \in \mathbb{R}^{g \times d}$, which are then subsequently used for cross-shape feature interaction and correspondence estimation. 

\begin{figure}[t]
    \centering
    \includegraphics[width=\linewidth]{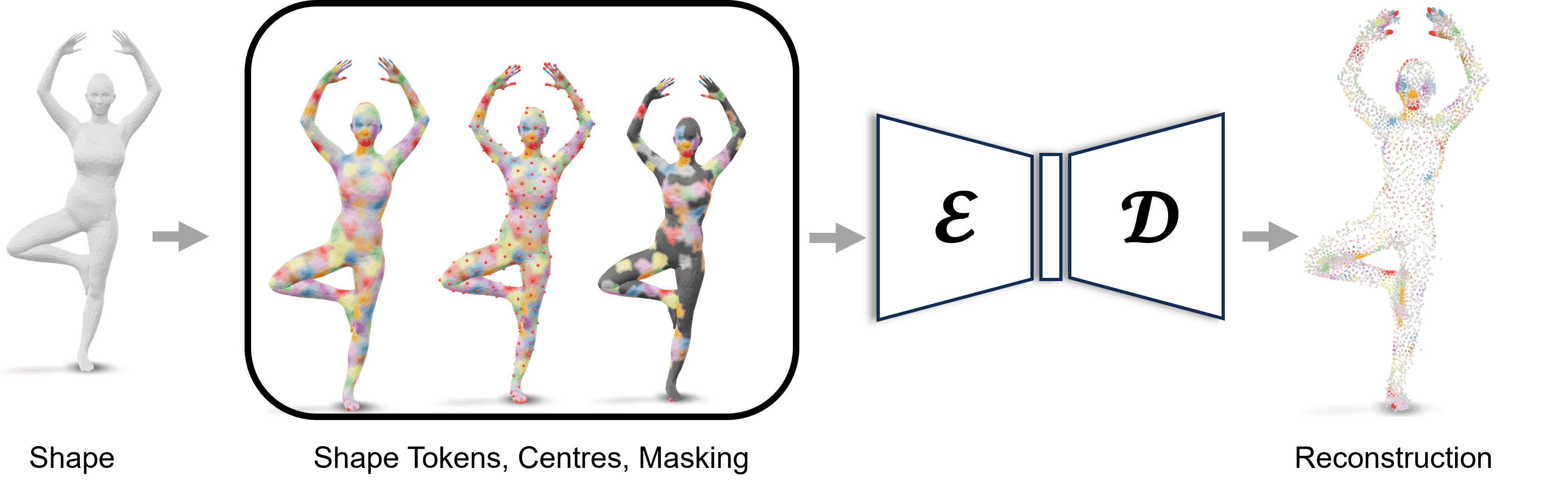}

    \caption{\textbf{Masked auto-encoding pre-training for mesh transformers.}
    A random subset of tokens is masked out given a geometry-aware tokenised mesh. The transformer encoder $\mathcal{E}$ processes only visible tokens, and the model (including the decoder $\mathcal{D}$) is trained to reconstruct the missing geometric information from the latent representation. This encourages learning of robust and semantically meaningful geometry-aware features that capture both local structure and global shape context. 
    }
    \label{fig:mae_pretraining}
\end{figure}

\paragraph{Masked auto-encoding pre-training.} 
To learn geometry-aware representations, we pre-train the mesh transformer using a masked auto-encoding (MAE) objective, following the design of MeshMAE~\cite{meshmae2022}; see Fig.~\ref{fig:mae_pretraining}. The pre-training encourages the model to recover missing geometric structures from partially observed token sequences, leading to robust and semantically meaningful representations. 
It simulates partiality during training and, thus, facilitates generalisation across partial-to-partial, full-to-full and partial-to-full matching scenarios.

Given the token embeddings $\mathbf{E} = \{\mathbf{e}_i\}_{i=1}^{g}$, we randomly sample a subset of token indices $\mathcal{I}_m \subset \{1,\dots,g\}$ with masking ratio $r$. Masked tokens are replaced by a learnable mask embedding $\mathbf{e}_{\text{mask}}$. The uncorrupted (visible) token sequence is processed by the transformer encoder. A lightweight decoder is then used to reconstruct the original geometric embeddings of the masked tokens.
\paragraph{Reconstruction loss.} 
During the self-supervised MAE pre-training, we supervise reconstruction with geometric and feature-level objectives. The Chamfer Distance (CD) loss $\mathcal{L}_{\text{CD}}$ is computed between the reconstructed and ground-truth point sets as follows: 
\begin{equation}
\mathcal{L}_{\text{CD}} = 
\frac{1}{|P|} \sum_{p \in P} \min_{q \in Q} \|p - q\|_2 +
\frac{1}{|Q|} \sum_{q \in Q} \min_{p \in P} \|q - p\|_2,
\end{equation}
where $P$ and $Q$ denote reconstructed and ground-truth surface samples, respectively.
Additionally, we apply a feature reconstruction loss $\mathcal{L}_{\text{feat}}$ on token embeddings for masked patches:
\begin{equation}
\mathcal{L}_{\text{feat}} = \frac{1}{g} \sum_{i=1}^{g} \|\hat{\mathbf{e}}_i - \mathbf{e}_i\|_2^2, 
\end{equation}
where $\hat{\mathbf{e}}_i$ and $\mathbf{e}_i$ are, respectively, the reconstructed and original token embeddings from the encoder. 
Our final training objective for the transformer backbone reads as follows: 
\begin{equation}
\mathcal{L}_{\text{MAE}} = \mathcal{L}_{\text{feat}} + \lambda \mathcal{L}_{\text{CD}}.
\end{equation}
with the weight $\lambda = 0.5$. 
After pre-training, we discard the decoder and retain only the encoder, as the decoder is required only for the reconstruction objective and not for correspondence estimation. 

\subsection{Cross-Shape Feature Interaction Learning}\label{sec:cross-attn} 
The encoded representations $\mathbf{Z}_{\mathcal{M}} \in \mathbb{R}^{g_{\mathcal{M}} \times d}$ and
$\mathbf{Z}_{\mathcal{N}} \in \mathbb{R}^{g_{\mathcal{N}} \times d}$ capture rich intra-shape geometric context through self-attention. 
These features are computed independently for each shape. 
To enable correspondence-aware reasoning, we introduce a cross-attention module that allows information exchange between the two shapes. Specifically, we update the features of $\mathcal{M}$ by attending to $\mathcal{N}$, and symmetrically for $\mathcal{N}$, as follows: 

\begin{equation}
\widetilde{\mathbf{Z}}_{\mathcal{M}} =
\mathrm{softmax}\!\left(\frac{\mathbf{Q}_{\mathcal{M}} \mathbf{K}_{\mathcal{N}}^\top}{\sqrt{d}}\right)\mathbf{V}_{\mathcal{N}}, \quad
\widetilde{\mathbf{Z}}_{\mathcal{N}} =
\mathrm{softmax}\!\left(\frac{\mathbf{Q}_{\mathcal{N}} \mathbf{K}_{\mathcal{M}}^\top}{\sqrt{d}}\right)\mathbf{V}_{\mathcal{M}},
\end{equation}

where $\mathbf{Q}_{\mathcal{M}} = \mathbf{Z}_{\mathcal{M}}\mathbf{W}_Q$,
$\mathbf{K}_{\mathcal{N}} = \mathbf{Z}_{\mathcal{N}}\mathbf{W}_K$, and
$\mathbf{V}_{\mathcal{N}} = \mathbf{Z}_{\mathcal{N}}\mathbf{W}_V$ are learned projections. 
This bidirectional cross-attention mechanism enables each token on one shape to aggregate geometry-aware contextual information from the other shape, \textit{effectively conditioning both representations on potential correspondences.} 
The resulting features $\widetilde{\mathbf{Z}}_{\mathcal{M}}$ and $\widetilde{\mathbf{Z}}_{\mathcal{N}}$ encode both intra-shape structure and inter-shape alignment cues, and are, therefore, used for subsequent functional map estimation. 

\textbf{PointInfoNCE loss.} To further improve correspondence-aware representation learning, we formulate a contrastive objective for point features. 
The loss $\mathcal{L}_{\mathrm{nce}}$ encourages features of corresponding points to be close in the embedding space, while non-corresponding points are pushed apart. 
We adopt the PointInfoNCE formulation of Attaiki et al.~\cite{ attaiki2021dpfm} and Xie et al.~\cite{xie2025echomatch}, which reads as follows: 
\begin{equation}
\mathcal{L}_{\mathrm{nce}}(\widetilde{\mathbf{Z}}_{\mathcal{M}}, \widetilde{\mathbf{Z}}_{\mathcal{N}}) = -\sum_{(i,j)\in\mathcal{P}} \log \frac{\exp(\widetilde{\mathbf{Z}}_{\mathcal{M}}(i) \cdot \widetilde{\mathbf{Z}}_{\mathcal{N}}(j) / \tau)}{\sum_{(\cdot,k)\in\mathcal{P}} \exp(\widetilde{\mathbf{Z}}_{\mathcal{M}}(i) \cdot \widetilde{\mathbf{Z}}_{\mathcal{N}}(k) / \tau)}. 
\end{equation}
$\mathcal{P}$ is the set of matched points (ground-truth correspondences), $\tau=0.07$ is the temperature parameter, and $\widetilde{\mathbf{Z}}{\mathcal{M}}(i)$ is the feature vector of point $i$ extracted from the cross-attended representation $\widetilde{\mathbf{Z}}{\mathcal{M}}$.
\subsection{Shape Overlap Prediction}\label{sec:overlap} 
Inspired by EchoMatch~\cite{xie2025echomatch}, we incorporate an overlap prediction module to estimate the shared region between $\mathcal{M}$ and $\mathcal{N}$. In contrast to Xie et al.~\cite{xie2025echomatch}, which predicts overlap from independently extracted point features, our overlap prediction operates on transformer-based cross-attended features, explicitly allowing us to leverage cross-shape contextual interactions. 
Given cross-attended features $\widetilde{\mathbf{Z}}_{\mathcal{M}}$ and $\widetilde{\mathbf{Z}}_{\mathcal{N}}$, we compute directional row-stochastic correspondences:
\begin{equation}
\boldsymbol{\Pi}_{\mathcal{MN}} = \mathrm{Softmax}\!\left(\frac{\widetilde{\mathbf{Z}}_{\mathcal{M}} \widetilde{\mathbf{Z}}_{\mathcal{N}}^{\top}}{\tau}\right) \quad\text{and}\quad 
\boldsymbol{\Pi}_{\mathcal{NM}} = \mathrm{Softmax}\!\left(\frac{\widetilde{\mathbf{Z}}_{\mathcal{N}} \widetilde{\mathbf{Z}}_{\mathcal{M}}^{\top}}{\tau}\right),
\end{equation}
where $\tau = 0.1$ is a temperature parameter. 
Overlap consistency is then estimated via cycle composition as $\mathbf{P}_{\mathcal{M}} = \boldsymbol{\Pi}_{\mathcal{MN}} \boldsymbol{\Pi}_{\mathcal{NM}}$. The diagonal entries of $\mathbf{P}_{\mathcal{M}}$ measure self-return probability after a round trip through $\mathcal{N}$, which is aggregated (with local neighbourhood smoothing) into per-vertex overlap scores $\mathbf{p}_{\mathcal{M}} \in [0,1]^{v_{\mathcal{M}}}$ and $\mathbf{p}_{\mathcal{N}} \in [0,1]^{v_{\mathcal{N}}}$. %
We supervise overlap prediction using ground-truth binary masks $\mathbf{m}_{\mathcal{M}}$ and $\mathbf{m}_{\mathcal{N}}$:
\begin{equation}
\mathcal{L}_{\mathrm{ov}} = \mathrm{BCE}(\mathbf{m}_{\mathcal{M}}, \mathbf{p}_{\mathcal{M}}) + \mathrm{BCE}(\mathbf{m}_{\mathcal{N}}, \mathbf{p}_{\mathcal{N}}), 
\end{equation}
where ``$\mathrm{BCE(\cdot, \cdot)}$'' denotes the binary cross-entropy function. 

\subsection{Functional Map Generation}\label{sec:fm}
\textbf{Partial functional maps.}
When $\mathcal{M}$ and $\mathcal{N}$ only overlap partially, directly solving Eq.~(3) can yield inaccurate correspondences due to mismatched descriptor support~\cite{attaiki2021dpfm, bracha2024unsupervised}. Following prior work~\cite{xie2025echomatch}, we restrict descriptors on $\mathcal{N}$ to the overlapping region using a binary mask $\mathbf{m}_{\mathcal{N}} \in \{0,1\}^{v_{\mathcal{N}}}$, i.e., $\tilde{\mathbf{F}}_{\mathcal{N}} = \mathbf{m}_{\mathcal{N}} \odot \mathbf{F}_{\mathcal{N}}$, where ``$\odot$'' denotes element-wise multiplication. We then compute $\mathbf{A}_{\mathcal{M}} = \boldsymbol{\Phi}_{\mathcal{M}}^{\dagger} \mathbf{F}_{\mathcal{M}}$ and $\tilde{\mathbf{A}}_{\mathcal{N}} = \boldsymbol{\Phi}_{\mathcal{N}}^{\dagger} \tilde{\mathbf{F}}_{\mathcal{N}}$, and estimate $\mathbf{C}_{\mathcal{M}\mathcal{N}}$ by replacing $\mathbf{A}_{\mathcal{N}}$ with $\tilde{\mathbf{A}}_{\mathcal{N}}$ in Eq.~(3).  
Masking only $\mathcal{N}$ avoids rank-deficient systems under small overlap. 

\textbf{Functional map loss.}  
We penalise the discrepancy between the predicted map $\mathbf{C}_{\mathcal{M}\mathcal{N}}$ and the ground-truth functional map $\mathbf{C}_{\mathcal{M}\mathcal{N}}^{\mathrm{gt}}$. The functional map loss reads: 
\begin{equation}
\mathcal{L}_{\mathrm{fmap}} = \left\| \mathbf{C}_{\mathcal{M}\mathcal{N}} - \mathbf{C}_{\mathcal{M}\mathcal{N}}^{\mathrm{gt}} \right\|_F^2. 
\end{equation}

\subsection{Final Training Objective}\label{ssec:final_training} 
To summarise, $\mathcal{L}_{\text{MAE}}$ is used during MAE pre-training. 
After that, we proceed with supervised training of TokenMatch, jointly optimising for the correspondences, overlaps, and feature representations using a combination of equally-weighted $\mathcal{L}_{\mathrm{fmap}}$, $\mathcal{L}_{\mathrm{ov}}$ and $\mathcal{L}_{\mathrm{nce}}$, resulting in the total objective: 
\begin{equation}
\mathcal{L}_{\mathrm{total}}
=
\mathcal{L}_{\mathrm{fmap}}
+
\mathcal{L}_{\mathrm{ov}}
+
\mathcal{L}_{\mathrm{nce}}.
\end{equation}

\begin{figure}
  \centering
  \includegraphics[width=\linewidth]{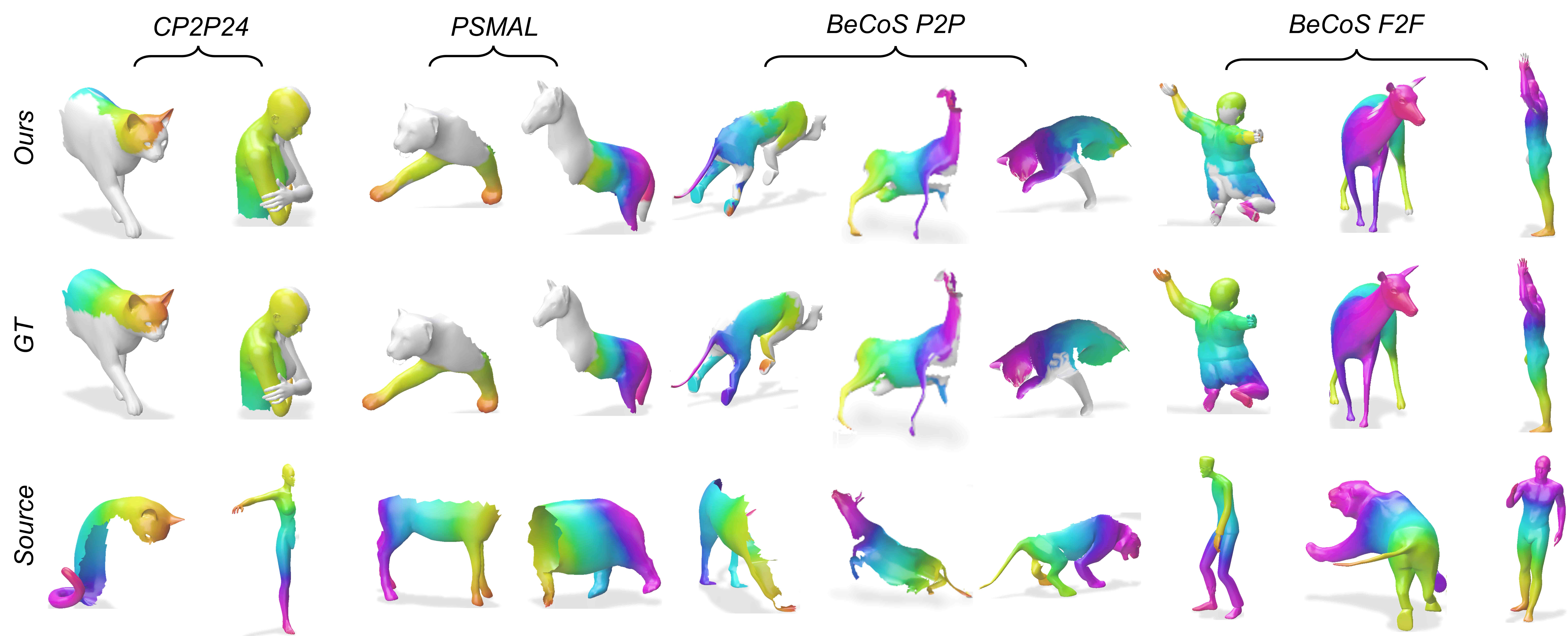}
  \caption{\textbf{Qualitative results on CP2P24, PSMAL and BeCoS datasets} 
  under varying levels of partiality and degree of non-isometric deformation. 
  Matches are colour-coded \textit{columnwise}. 
  The full-full (\textit{F2F}) setting shows results for the model trained on partial-partial (\textit{P2P}). 
  These examples offer a balanced visualisation of our overall results by TokenMatch. 
  ``GT'' denotes ground truth. 
   }
  \label{fig:qualitative}
\end{figure}

\section{Experimental Evaluation}\label{sec:experiments} 
This section presents a comprehensive evaluation of TokenMatch on both partial and full shape correspondence estimation tasks. We begin by describing the datasets, baseline methods, and evaluation metrics used throughout our experiments (Sec.~\ref{ssec:datasets_baslines_metrics}). We then provide implementation details (Sec.~\ref{sec:implementation}) and report quantitative and qualitative results on challenging partial-to-partial matching benchmarks, followed by standard full shape correspondence benchmarks to assess performance under complete geometry (Sec.~\ref{ssec:experiments}). Finally, we analyse the robustness and generalisation of our approach under more challenging settings, including anisotropic remeshing and cross-category transfer, highlighting the stability of TokenMatch across varying shape representations and deformation regimes. 

\subsection{Datasets, Baselines and Evaluation Metrics}\label{ssec:datasets_baslines_metrics} 

\textbf{Partial-shape matching datasets.}
We evaluate robustness under partiality using all major benchmarks for partial-to-partial shape correspondence, spanning both near-isometric and strongly non-isometric regimes. We consider (i) the CP2P (Cuts-Partial-to-Partial)~\cite{attaiki2021dpfm, ehm2024partial} dataset, which is derived from SHREC’16~\cite{cosmo2016shrec} and TOSCA~\cite{bronstein2008numerical} and which contains scale-normalised, pre-aligned partial shapes generated via simulated cuts and primarily evaluates correspondence under controlled, near-isometric conditions across humanoid and animal models; (ii) the PSMAL (PARTIAL-SMAL)~\cite{ehm2024partial} dataset derived from SMAL~\cite{zuffi20173d}, with substantial non-isometric deformations in animal shapes while maintaining normalised scale, thereby increasing geometric variability and task difficulty; and (iii) the BeCoS~\cite{ehm2025beyond} dataset, a significantly more challenging benchmark than (i) and (ii), which aggregates partial humanoid and animal meshes from diverse datasets~\cite{anguelov2005scape, bogo2014faust, bronstein2008numerical, dyke2020shrec, magnet2022smooth, rodola2014dense, zuffi20173d} and incorporates realistic scaling, strong non-isometric deformations, noise, and varying levels of incompleteness, reflecting more practical and unconstrained shape matching scenarios. 

\begin{table}[!t]
\centering
\caption{\textbf{Mean IoU (×100) on different partial-to-partial shape matching datasets: CP2P, PSMAL and BeCoS.} Prior methods rely on predefined descriptors (XYZ \cite{ehm2025beyond} or DINOv2 \cite{oquab2023dinov2}), while our approach learns features directly from mesh geometry. The best results are shown in \textbf{bold}.} 
\label{tab:pp_results}
\vspace{5pt}
\begin{tabular}{llccc}
\toprule
\textbf{Method} & \textbf{Feature Type} & \textbf{CP2P24} $\uparrow$ & \textbf{PSMAL} $\uparrow$ & \textbf{BeCoS} $\uparrow$ \\
\midrule

\multirow{2}{*}{SM-COMB~\cite{roetzer2022scalable}} 
 & XYZ    & 57.86 & 54.76 & 47.04 \\
 & DINOv2 & 38.38 & 36.61 & 48.29 \\

\multirow{2}{*}{GC-PPSM~\cite{ehm2024partial}} 
 & XYZ    & 69.29 & 64.34 & 49.34 \\
 & DINOv2 & 49.66 & 34.30 & 33.14 \\

\multirow{2}{*}{DPFM~\cite{attaiki2021dpfm}} 
 & XYZ    & 63.86 & 67.04 & 48.18 \\
 & DINOv2 & 74.15 & 73.67 & 51.02 \\

\multirow{2}{*}{EchoMatch~\cite{xie2025echomatch}} 
 & XYZ    & 80.10 & 72.71 & 52.40 \\
 & DINOv2 & 84.72 & 84.75 & 64.68 \\

\midrule

\multirow{1}{*}{\textbf{TokenMatch (Ours)}} 
 & Learned (mesh) & \textbf{85.56} & \textbf{85.21} & \textbf{65.25} \\ 

\bottomrule
\end{tabular}
\vspace{-15pt}
\end{table}

\begin{wraptable}{r}{0.6\textwidth} %
\vspace{-8pt}
  \centering
  \caption{\textbf{Mean geodesic error ($\times 100$) on FAUST, SCAPE, and SHREC’19.} The best results are shown in \textbf{bold}.}
  \label{tab:main_results}

  \vspace{-4pt}
  \setlength{\tabcolsep}{4pt}
  \renewcommand{\arraystretch}{1.05}
  \small

  \begin{tabular}{lccc}
    \toprule
    \textbf{Method} & \textbf{FAUST} $\downarrow$ & \textbf{SCAPE} $\downarrow$ & \textbf{SHREC’19} $\downarrow$ \\
    \midrule
    3D-CODED~\cite{groueix20183d}        & 2.50 & 16.10 & 17.30 \\
    TransMatch~\cite{trappolini2021shape} & 1.70 & 15.30 & 21.00 \\
    DUO-FMNet~\cite{donati2022deep}       & 2.50 & 4.20  & 6.40 \\
    GeomFMaps~\cite{donati2020deep}       & 1.90 & 2.40  & 7.90 \\
    AttentiveFMaps~\cite{li2022learning}  & 1.90 & 2.60  & 5.80 \\
    ConsistentFMaps~\cite{sun2023spatially} & 2.30 & 2.60 & 3.80 \\
    SSL~\cite{cao2023self}               & 2.00 & 3.10  & 4.00 \\
    DiffZO~\cite{magnet2024memory}        & 1.90 & 2.40  & 4.20 \\
    ULRSSM~\cite{cao2023robust}           & 1.60 & 2.20  & 5.70 \\
    SmS~\cite{cao2024spectral}            & \textbf{1.40} & 3.30 & 6.20 \\
    DenoisFM~\cite{zhuravlev2025denoising} & 1.70 & 2.10 & 3.90 \\
    \midrule
    \textbf{TokenMatch (Ours)} & 1.72 & \textbf{2.09} & \textbf{3.45} \\
    \bottomrule
  \end{tabular}
\end{wraptable}

\textbf{Full-shape matching datasets.} 
We evaluate TokenMatch on three widely used benchmarks for full shape matching: (i) the FAUST dataset~\cite{bogo2014faust} of human meshes in pose variations; (ii) the SCAPE dataset~\cite{anguelov2005scape}, consisting of a single subject in diverse poses; and (iii) the SHREC’19 dataset~\cite{melzi2019shrec}, which includes human shapes with substantial variation in body proportions and pose, making it particularly challenging. SHREC’19 is used solely for testing, and we exclude one outlier shape following common practice to keep results comparable~\cite{cao2023robust}. We fine-tune a partial-shape model for~\Cref{tab:main_results} and~\Cref{tab:anisotropic_results}; all other experiments use a model trained only on BeCoS partial-to-partial data. 
Performance on the standard FAUST and SCAPE benchmarks has largely saturated \cite{zhuravlev2026nonrigid}. 
We include them for completeness, as our focus is on more challenging correspondence estimation scenarios and cross-setting generalisability. 
\textbf{Baselines.} 
Despite its high practical relevance, partial-to-partial shape matching remains under-explored. 
Existing approaches for this setting include learning-based (EchoMatch~\cite{xie2025echomatch} and DPFM~\cite{attaiki2021dpfm}) and axiomatic techniques (SM-COMB~\cite{roetzer2022scalable} and GC-PPSM~\cite{ehm2024partial}). 
In addition, we consider recent methods for non-rigid full shape correspondence, which can be broadly grouped into template-based learning and descriptor-based methods. 
The first category includes large-scale template-based models trained on synthetic datasets such as SURREAL~\cite{varol2017learning}, including 3D-CODED~\cite{groueix20183d} and TransMatch~\cite{trappolini2021shape}, which learn dense correspondences across shapes in a canonical template framework using full-shape supervision. 
The second category consists of descriptor-based approaches that compute correspondences from learned or handcrafted shape features, including ULRSSM~\cite{cao2023robust}, DiffZO~\cite{magnet2024memory}, ConsistentFMaps~\cite{sun2023spatially}, GeomFMaps~\cite{donati2020deep}, AttentiveFMaps~\cite{li2022learning}, DUO-FMNet~\cite{donati2022deep}, SSL~\cite{cao2023self}, and SmS~\cite{cao2024spectral}. 
These methods rely on full-shape training or precomputed shape descriptors. 

While all these methods are developed and evaluated in the full-shape setting, TokenMatch is trained exclusively on annotated partial-to-partial correspondences yet generalises to full-shape matching without additional training. 

\textbf{Evaluation metrics.}
Following established practices~\cite{xie2025echomatch, attaiki2021dpfm}, we report two evaluation measures: (1) \textit{mIoU (Mean Intersection over Union)}, which quantifies overlap prediction quality by computing the intersection over union between predicted and ground-truth binary masks over vertices, where higher values indicate better overlap estimation; and (2) \textit{GE (Geodesic Error)}~\cite{kim2011blended}, which measures correspondence accuracy. In the partial-to-partial setting, GE is computed using geodesic distances normalised by the square root of the full-shape area to ensure scale invariance. We report mean GE, which is evaluated over the predicted overlapping region. 

\subsection{Implementation Details}\label{sec:implementation}
For pre-training, following Liang et al.~\cite{meshmae2022}, we use ViT-Base, a transformer backbone with 12 layers, a hidden width of 768, an MLP dimension of 3072, and 12 attention heads, with $g = 256$ tokens used for all shapes. 
For overlap prediction, following Xie et al.~\cite{xie2025echomatch}, we use DiffusionNet~\cite{sharp2022diffusionnet} blocks with a hidden dimension of 16 and $k = 50$ Laplace--Beltrami eigenfunctions, such that the functional map $C \in \mathbb{R}^{50 \times 50}$. We initialise the soft correspondence temperature to $\tau = 0.1$ and treat it as a learnable parameter during training. Final point-to-point correspondences within the predicted overlap region are obtained via nearest-neighbour matching in the feature space.

We conduct all experiments on a single node with two NVIDIA A100 80GB PCIe GPUs. The implementation is based on CUDA 13.1 \cite{CUDA131}. No other processes are executed during training. 

\subsection{Experimental Results}\label{ssec:experiments}
We evaluate TokenMatch in partial-shape matching scenarios; see Table~\ref{tab:pp_results}. 
Our approach outperforms all baselines across CP2P~\cite{attaiki2021dpfm, ehm2024partial}, PSMAL~\cite{ehm2024partial}, and BeCoS~\cite{ehm2025beyond}. 
The improvements are pronounced on BeCoS, where strong geometric variability and missing regions make correspondence estimation highly challenging. 
Previous methods rely on predefined descriptors (XYZ and DINOv2~\cite{oquab2023dinov2}). 
Among them, EchoMatch~\cite{xie2025echomatch} achieves the second-best performance, whereas we learn geometry-aware representations and features directly from the mesh structure. 
This highlights the benefit of task-specific, geometry-native representations, together with masked pre-training, for robust correspondence learning under incomplete geometry across all benchmarks. 
\begin{wraptable}{r}{0.48\linewidth}
\centering
\small
\vspace{3pt}
\caption{Results on anisotropic FAUST and SCAPE datasets. 
\textbf{Bold} highlights best results.} 
\vspace{-5pt}
\label{tab:anisotropic_results}
\setlength{\tabcolsep}{4pt}
\begin{tabular}{lcc}
\toprule
\textbf{Model} & \textbf{FAUST\_a} $\downarrow$ & \textbf{SCAPE\_a} $\downarrow$ \\
\midrule
3D-CODED~\cite{groueix20183d}        & 2.90 & 16.90 \\
TransMatch~\cite{trappolini2021shape}      & 2.70 & 16.20 \\
DUO-FMNet~\cite{donati2022deep}       & 3.00 & 4.40 \\
GeomFMaps~\cite{donati2020deep}       & 3.20 & 3.80 \\
AttentiveFMaps~\cite{li2022learning}  & 2.40 & 2.80 \\
ConsistentFMaps~\cite{sun2023spatially} & 2.60 & 2.70 \\
ULRSSM~\cite{cao2023robust}          & \textbf{1.90} & 2.40 \\
DenoisFM~\cite{zhuravlev2025denoising}        & 2.20 & 2.20 \\
\midrule
\textbf{TokenMatch (Ours)}            & \textbf{1.90} & \textbf{1.70} \\
\bottomrule
\end{tabular}
\vspace{-10pt}
\end{wraptable}

Fig.~\ref{fig:qualitative} shows qualitative results under partial and full shape matching. 
Table~\ref{tab:main_results} reports results on FAUST~\cite{bogo2014faust}, SCAPE~\cite{anguelov2005scape}, and SHREC’19~\cite{melzi2019shrec}, i.e., standard benchmarks for full-shape matching. 
Our TokenMatch achieves consistently strong performance across all datasets, with particularly notable gains on the challenging (due to different mesh connectivities) SHREC'19~\cite{melzi2019shrec} benchmark. 
While some methods achieve competitive performance on individual datasets~\cite{pierson2025diffumatch, attaiki2021dpfm}, they often degrade under larger test distribution shifts. 
SmS~\cite{cao2024spectral} and ULRSSM~\cite{cao2023robust} rank best on FAUST. 
Our approach is competitive with TransMatch~\cite{trappolini2021shape} and DenoisFM~\cite{zhuravlev2025denoising} among the five most accurate methods, while also generalising from partial-to-partial training to full-shape matching without additional training. 
TokenMatch maintains stable performance across all settings, indicating improved generalisation to diverse non-rigid deformations. 
Additional ablation studies covering tokenisation design, geometric representation (mesh vs.~point cloud), cross-shape interaction, token overlap degree and number of mesh tokens are included in Apps.~\ref{subsec:token-design}--\ref{subsec:num-token}. 
Curvature guidance with soft token assignment, i.e., our main technical contribution, results in the strongest model. 
\textbf{Robustness to anisotropic remeshing.}
To assess the sensitivity of different methods to mesh connectivity, we evaluate our method on anisotropically remeshed versions of FAUST and SCAPE~\cite{anguelov2005scape, bogo2014faust}. 
These meshes exhibit non-uniform triangle sizes, with regions of both dense and coarse sampling, making them challenging for methods that depend on consistent connectivity or local neighbourhood structure \cite{groueix20183d, trappolini2021shape, donati2022deep}. 
The performance of most competing approaches degrades noticeably under these conditions, reflecting their sensitivity to irregular discretisation; see Table~\ref{tab:anisotropic_results}. 
In contrast, TokenMatch remains stable and consistently accurate across both standard and anisotropic meshing settings. 
We believe this robustness stems from curvature-guided tokenisation, which captures geometry independent of mesh resolution, and the transformer operating on token-based representations rather than explicit connectivity, yielding stable and high accuracy across varying mesh structures. 
\begin{wraptable}{r}{0.34\linewidth}
\vspace{-12pt}
\centering
\small
\setlength{\tabcolsep}{4pt}

\caption{Results on SMAL~\cite{zuffi20173d}. 
}
\label{tab:smal_results}

\begin{tabular}{lcccc}
\toprule
\textbf{Class} & \textbf{Ours} & \textbf{\cite{zhuravlev2025denoising}} & \textbf{\cite{cao2023robust}} & \textbf{\cite{sun2023spatially}} \\
\midrule
wolf  & \textbf{3.40} & 3.50 & 3.50 & 4.40 \\
dog   & \textbf{3.30} & 3.50 & 3.50 & 4.50 \\
horse & \textbf{3.60} & 3.70 & 3.80 & 4.40 \\
cow   & \textbf{3.30} & 3.80 & 3.80 & 4.40 \\
lion  & \textbf{3.60} & 4.00 & 3.80 & 4.90 \\
fox   & 4.10 & 4.70 & \textbf{3.60} & 4.90 \\
hippo & \textbf{7.40} & 7.60 & 9.90 & 7.90 \\
\midrule
Mean  & \textbf{4.10} & 4.30 & 4.30 & 4.80 \\
\bottomrule
\end{tabular}
\vspace{-12pt}
\end{wraptable}

\begin{figure}[t!]
  \centering
  \includegraphics[width=\linewidth]{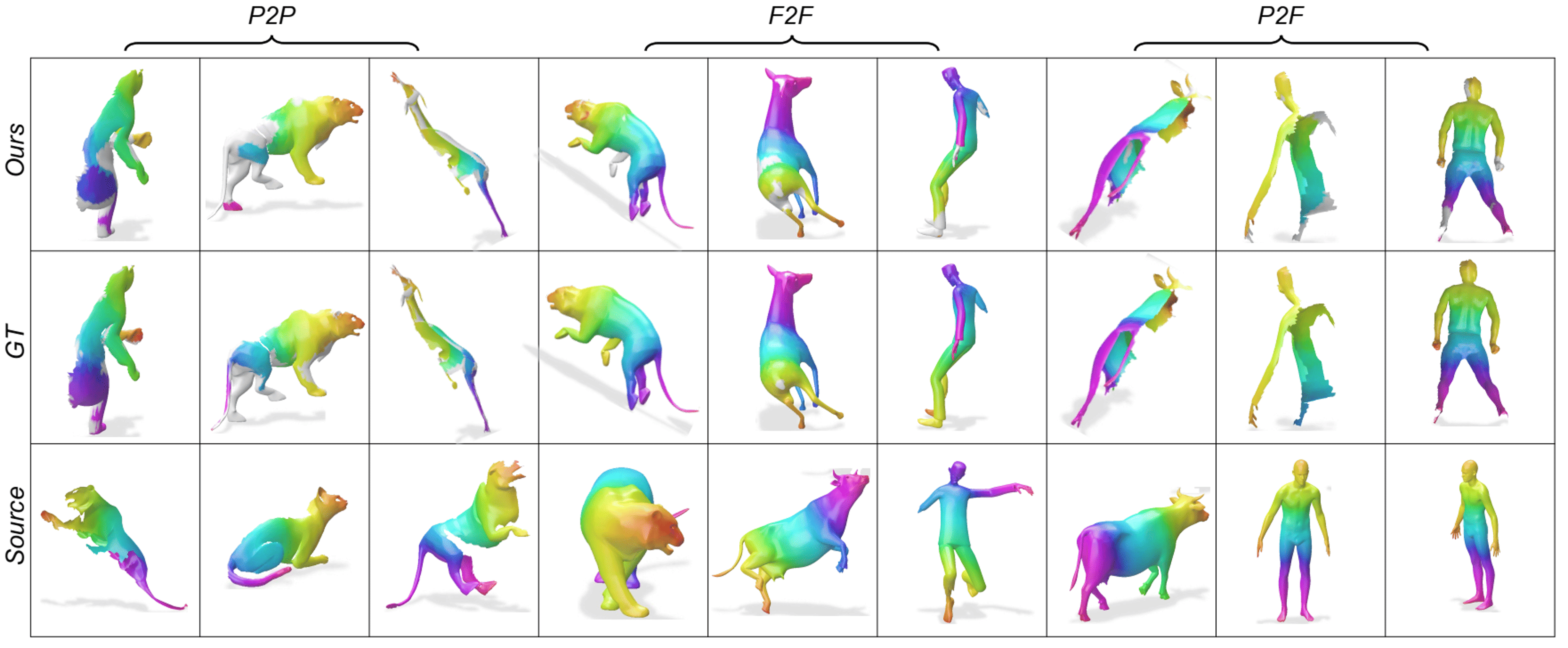}
  \caption{\textbf{Qualitative results on the BeCoS dataset \cite{ehm2025beyond}.} Correspondences are shown under varying levels of partiality and deformation, with matches colour-coded \textit{columnwise}. Notably, despite being trained exclusively on partial-to-partial (\textit{P2P}) correspondences, the model generalises effectively to both full-to-full (\textit{F2F}) and partial-to-full (\textit{P2F}) settings. ``GT'' denotes ground truth. 
  }
  \label{fig:qualitative2}
  \vspace{-5pt}
\end{figure}

\vspace{5pt}
\textbf{Generalisation across categories.} 
Table~\ref{tab:smal_results} reports results on SMAL~\cite{zuffi20173d} animal categories. 
Our model achieves the lowest mean GE across all but one category, including challenging cases such as \textit{hippo}, \textit{horse} and \textit{lion}, where shape variability is high and local geometry is less consistent than in simpler shapes. 
Moreover, we evaluate a model (trained on the BeCoS~\cite{ehm2025beyond} partial-to-partial shapes) on the BeCoS full-to-full scenario; see Figs.~\ref{fig:qualitative} and~\ref{fig:qualitative2} for qualitative results. Table~\ref{tab:partial_full_results} and Fig.~\ref{fig:qualitative2} demonstrate the strong generalisability of TokenMatch: Despite being trained exclusively on partial-to-partial correspondences, the model successfully handles full-to-full and partial-to-full matching while maintaining consistently accurate correspondences under severe partiality and deformation. 
There are only a few artefacts in the transferred colours. 
{
\setlength{\columnsep}{12pt} %

\begin{wraptable}{r}{0.46\linewidth}
\centering
\scriptsize
\setlength{\tabcolsep}{3pt}
\vspace{-10pt}
\caption{Mean IoU results under different training and testing settings on the BeCoS dataset.}
\label{tab:partial_full_results}

\begin{tabularx}{\linewidth}{
    @{}
    >{\raggedright\arraybackslash}X
    >{\raggedright\arraybackslash}X
    >{\centering\arraybackslash}p{1.45cm}
    @{}
}
\toprule
\makecell[cl]{\textbf{Setting}} &
\makecell[cl]{\textbf{Protocol}} &
\makecell[cc]{\textbf{Mean IoU}\\ \textbf{($\times 100$) $\uparrow$}} \\
\midrule
Partial-to-Partial & Trained and Tested & 65.25 \\
Partial-to-Full    & Tested only        & 55.61 \\
Full-to-Full       & Tested only        & \textbf{71.85} \\
\bottomrule
\end{tabularx}

\vspace{-12pt}
\end{wraptable}
}

\vspace{5pt}
\textbf{Training and inference time.} Our method requires $\approx$six hours on BeCoS (partial-to-partial) for training and $0.16$ seconds per shape pair at inference (overlap prediction, functional map recovery and nearest-neighbour recovery for point matches), measured on a single node with two NVIDIA A100 80GB PCIe GPUs. 
Table~\ref{tab:runtime_comparison} shows that TokenMatch is comparable in training cost and faster at inference than DPFM and EchoMatch. It is also substantially faster than optimisation-based methods \cite{ehm2024partial, roetzer2022scalable}. 

\textbf{Masked pre-training and robustness to noise.}
As we show in App.~\ref{ssec:masking_ratio}, masked pre-training further enhances partial shape matching by improving the model’s ability to reason under missing or incomplete geometric observations. 
In App.~\ref{ssec:noise_robustness}, we expose our method to different levels of added noise during inference to assess its robustness. 
We find that TokenMatch remains robust to moderate levels of noise without noise augmentation during training. 

\setlength{\columnsep}{12pt} %

\begin{wraptable}{r}{0.46\linewidth}
\vspace{-12pt}
\centering
\scriptsize
\setlength{\tabcolsep}{3pt}

\caption{Training and inference times.}
\label{tab:runtime_comparison}

\begin{tabularx}{\linewidth}{
    @{}
    >{\raggedright\arraybackslash}X
    >{\centering\arraybackslash}p{1.55cm}
    >{\centering\arraybackslash}p{1.8cm}
    @{}
}
\toprule
\multicolumn{1}{@{}l}{\makecell[cl]{\textbf{Method}}} &
\multicolumn{1}{c}{\makecell[cc]{\textbf{Training}\\ \textbf{(whole set)}}} &
\multicolumn{1}{c@{}}{\makecell[cc]{\textbf{Inference}\\ \textbf{(per shape pair)}}} \\
\midrule
SM-COMB~\cite{roetzer2022scalable}   & N/A   & 0.1 h  \\
GC-PPSM~\cite{ehm2024partial}   & N/A   & 3.1 h  \\
DPFM~\cite{attaiki2021dpfm}      & 4.8 h & 0.18 s \\
EchoMatch~\cite{xie2025echomatch} & 5.4 h & 0.20 s \\
Ours      & 6 h   & \textbf{0.16 s} \\
\bottomrule
\end{tabularx}

\vspace{-8pt}
\end{wraptable}

\newpage
\section{Conclusion}\label{sec:conclusion}
We demonstrate a generalisable and unified model, serving as a foundation for partial and full shape-matching tasks. 
Our extensive experiments show consistent improvements over prior methods across standard benchmarks, along with strong robustness of our TokenMatch approach to partiality, non-rigid (non-isometric) deformations, and varying geometric complexity. 
Global reasoning enabled by transformer attention, when coupled with functional map supervision, consistently outperforms local descriptor-based approaches, particularly in challenging partial correspondence scenarios. 
This paper also demonstrates that geometry-native representations exhibit stronger transferability than fixed or generic pre-trained features such as raw XYZ embeddings or DINOv2 features, emphasising the importance of task-aligned geometric encoding. 
We experimentally validate our main technical contribution, i.e., 
curvature-guided mesh tokenisation with patch overlapping, which is necessary to achieve the reported accuracy of TokenMatch. 
Notably, training exclusively on partial correspondences is sufficient to achieve strong full-shape matching performance at inference time, without architectural modification or retraining. 
This effectively unifies full-to-full matching as a special case of partial-to-partial correspondence estimation, thereby reducing reliance on complete shape datasets. 
Along with its generalisability, TokenMatch is computationally efficient, requiring a fraction of a second per shape pair while maintaining high matching accuracy. 
\noindent\textbf{Future Work.} The proposed TokenMatch approach relies on geodesic computations, which can become expensive at high mesh resolutions (e.g., higher than what is common in the current datasets). 
We include qualitative examples of observed inaccuracies: Attentive readers can recognise them with the help of the transferred colours in Figs.~\ref{fig:qualitative} and \ref{fig:qualitative2}, as well as Figs.~\ref{fig:qualadd1} and  \ref{fig:becos_generalization} in the Appendix. 
Future work will explore scalable alternatives to geodesic processing, improve robustness to noisy, unstructured geometry and extend the framework to large-scale real-world scan collections. 
We hope this work will open up new perspectives towards \textit{(i)} a 
fully-fledged foundation model for shape correspondence estimation and towards \textit{(ii)} addressing long-standing challenges of this research field. 

\vspace{5pt} 

{
\small
\bibliographystyle{abbrv}
\bibliography{main}

}

\clearpage

\appendix

\renewcommand{\theHsection}{A.\arabic{section}}
\let\titleold\title
\renewcommand{\title}[1]{\titleold{#1}\newcommand{\thetitle}{#1}}

\newcommand{\maketitlesupplementary}{
   \newpage
   \begin{center}
       \Large
       \textbf{\thetitle} \\
       \vspace{0.5em}
       {\large Supplementary Material} \\
       \vspace{2.0em}
   \end{center}
}

\title{TokenMatch: 3D Mesh Correspondence Transformer \\with Curvature-Guided Tokenisation}

\maketitlesupplementary

\suppressfloats[t]

\noindent This Appendix provides additional technical details and experimental results. 
App.~\ref{sec:splits} outlines dataset preprocessing and evaluation protocols across all benchmarks used in both partial and full-shape matching settings. 

In App.~\ref{sect:ablation}, we provide an in-depth analysis of our mesh tokenisation design space (App.~\ref{subsec:token-design}) and an ablation study (variant without cross-attention) (App.~\ref{subsec:cross-atten}) that motivate our curvature-guided overlapping formulation. We further investigate the sensitivity of key hyperparameters, such as degree of token overlap (App.~\ref{subsec:overlap-degree}), the number of tokens (App.~\ref{subsec:num-token}), and masking ratio (App.~\ref{ssec:masking_ratio}). 
We also evaluate the robustness and accuracy of TokenMatch to increasingly noisy inputs (App.~\ref{ssec:noise_robustness}). 
Finally, App.~\ref{sec:additional_qualitative} provides additional qualitative results using the CP2P24 protocol and PSMAL dataset \cite{ehm2024partial} to test our method under varying levels of partiality and non-isometric deformation. These visualisations show that TokenMatch produces accurate and geometrically consistent correspondences even in challenging partial-to-partial settings with limited overlap and incomplete local structure. We also report more results on BeCoS, which features realistic scenes with noise, missing regions, and strong deformations. 
Our model accurately matches full shapes across different degrees of partiality, with stable correspondences preserving semantic alignment under large geometric variations. 

\section{Dataset Split Details}\label{sec:splits}
We evaluate our method on both partial-to-partial and full shape matching tasks. For partial matching, we use three benchmark datasets: CP2P~\cite{attaiki2021dpfm, ehm2024partial}, PSMAL~\cite{ehm2024partial}, and BeCoS~\cite{ehm2025beyond}. For full shape matching, we additionally consider the complete shape settings provided by full-to-full shape datasets. Below, we detail the dataset splits used in our experiments. 

\textbf{CP2P.} The CP2P (Cuts-Partial-to-Partial) benchmark was introduced in Attaiki et al.~\cite{attaiki2021dpfm} and is based on the SHREC16~\cite{cosmo2016shrec} and TOSCA~\cite{bronstein2008numerical} datasets. In our main experiments, we follow the CP2P24 protocol~\cite{attaiki2021dpfm, ehm2024partial}, which uses 120 shapes from the SHREC16~ CUTS training set and evaluates on 100 test pairs sampled from 153 shapes in the CUTS24 test split of SHREC16. 

\textbf{PSMAL.} The PSMAL dataset~\cite{ehm2024geometrically} is derived from SMAL~\cite{zuffi20173d} and consists of normalised non-isometric partial animal shapes across eight categories. We follow the standard evaluation protocol, which defines species-disjoint training and test splits over 49 shapes, resulting in 273 training pairs and 100 test pairs.

\textbf{BeCoS.} The large-scale BeCoS benchmark~\cite{ehm2025beyond} contains realistically scaled non-isometric partial shapes of humans and animals. It provides 10,185 training instances, 137 validation instances, and 142 test instances.

\textbf{FAUST.} The FAUST dataset~\cite{bogo2014faust} consists of 100 watertight human meshes covering ten identities in ten poses. Following the standard remeshing protocol~\cite{donati2020deep}, an 80/20 split is used for training and testing, respectively.

\textbf{SCAPE.} The SCAPE dataset~\cite{anguelov2005scape} contains 71 meshes of a single subject in different poses. The standard evaluation uses the first 51 shapes for training and the remaining 20 shapes for testing.

\textbf{SHREC'19.} The SHREC'19 benchmark~\cite{melzi2019shrec} includes 44 shapes with diverse body types and poses, and is commonly used as a test-only setting. No training split is defined, and evaluation is performed on all available shapes except shape 40, which is excluded because it is non-watertight. 

\begin{figure*}[!h]
    \centering
    \setlength{\tabcolsep}{2pt} %

    \begin{subfigure}[t]{0.23\textwidth}
    \centering
    \includegraphics[height=7cm]{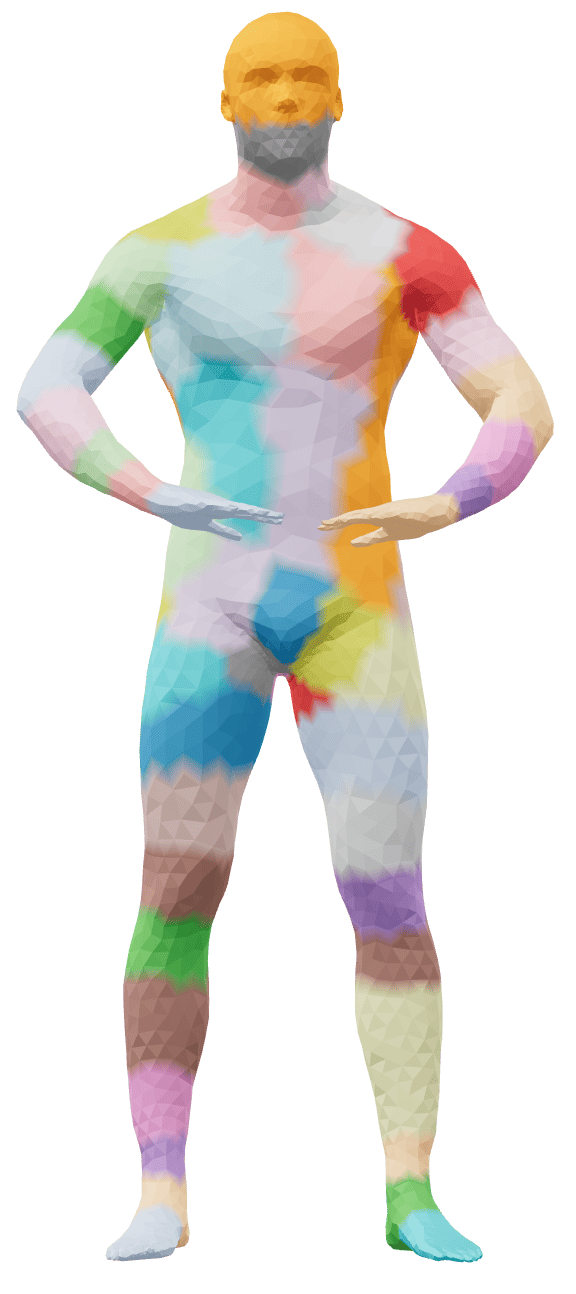}
    \caption{Spectral Laplacian}
\end{subfigure}
\hspace{0.2em}
\begin{subfigure}[t]{0.23\textwidth}
    \centering
    \includegraphics[height=7cm]{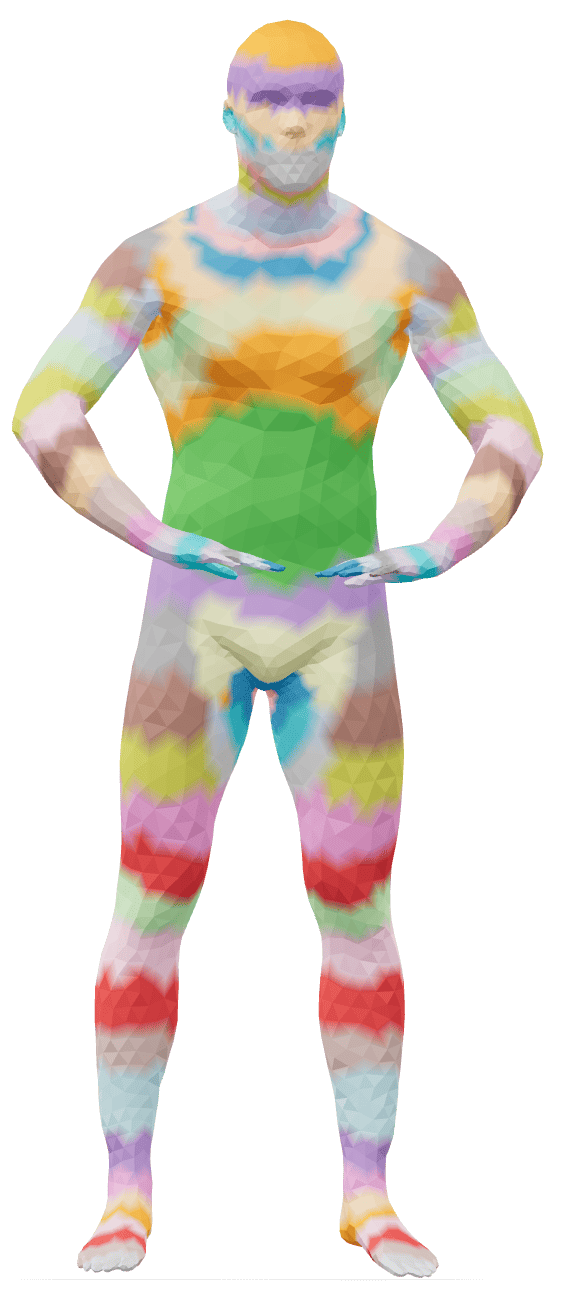}
    \caption{HKS clustering}
\end{subfigure}
\hspace{0.2em}
\begin{subfigure}[t]{0.23\textwidth}
    \centering
    \includegraphics[height=7cm]{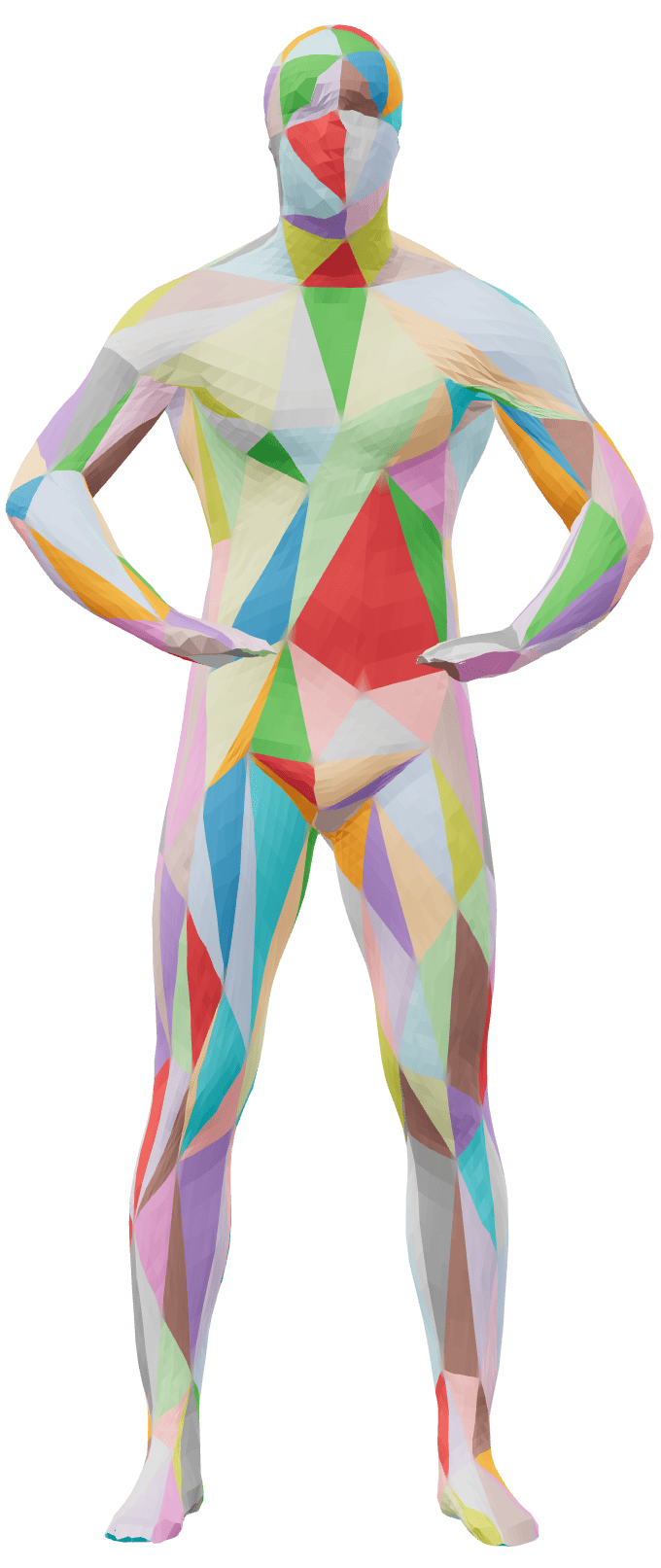}
    \caption{Face splitting}
\end{subfigure}
\hspace{0.2em}
\begin{subfigure}[t]{0.23\textwidth}
    \centering
    \includegraphics[height=7cm]{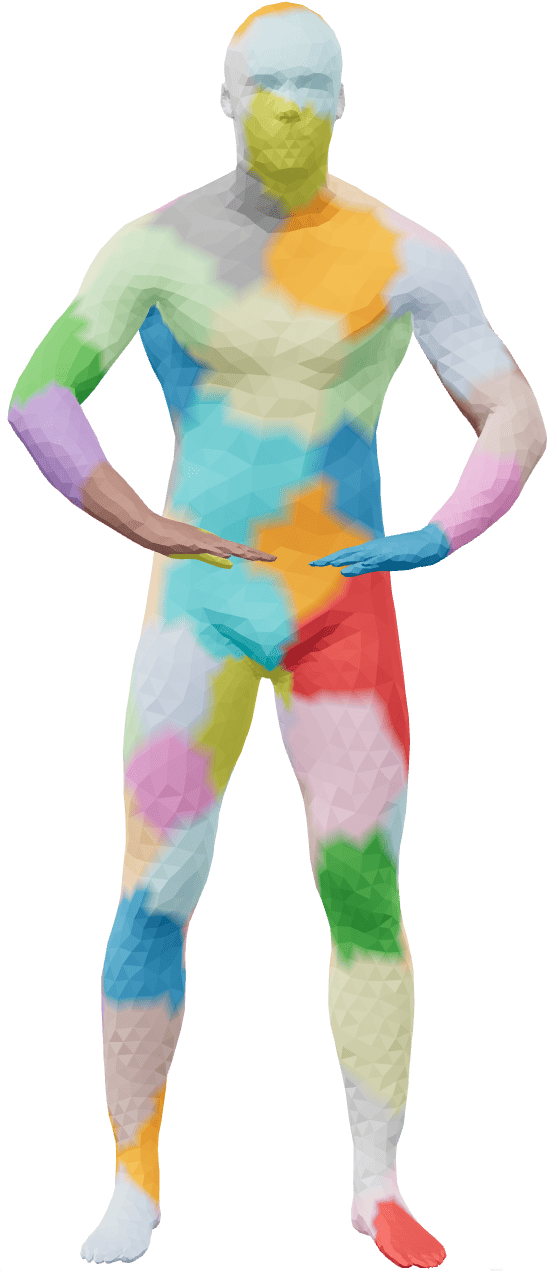}
    \caption{Voronoi (FPS)}
\end{subfigure}
    \caption{ %
    \textbf{Tokenisation design exploration.} 
    We compare alternative mesh tokenisation strategies. Spectral clustering groups vertices in a global embedding space, HKS uses diffusion-based geometric descriptors, face-based splitting enforces local structure via hierarchical subdivision, and Voronoi partitioning provides uniform geodesic coverage. These highlight trade-offs between global structure, locality, and spatial continuity, motivating our curvature-guided overlapping tokenisation. 
    }
    \label{fig:tokenization_comparison}
\end{figure*}

\section{Ablation Studies}\label{sect:ablation}
\subsection{Tokenisation Design Exploration}\label{subsec:token-design}

While the proposed TokenMatch relies on curvature-guided tokenisation, we explore several alternative mesh tokenisation strategies before arriving at our final curvature-guided overlapping design, each revealing different trade-offs between global structure, local geometry, and spatial continuity. We next refer to the elements of Fig.~\ref{fig:tokenization_comparison}. 
\textit{Spectral Laplacian clustering}~\cite{belkin2001laplacian} (a) computes Laplace--Beltrami eigenvectors and embeds each vertex into spectral space, where K-means is applied to form patches. While effective at capturing global geometric structure, it often produces spatially disconnected regions. \textit{Heat Kernel Signature (HKS) clustering}~\cite{bronstein2010scale} (b) replaces spectral embeddings with multi-scale HKS descriptors, yielding more deformation-robust and geometry-sensitive patches, but tends to produce overly diffuse regions with weak boundary adherence. \textit{Face-based hierarchical splitting}~\cite{meshmae2022} (c) recursively subdivides mesh faces in a top-down manner, representing each patch via local geometric features such as area, angles, and normals. This yields highly structured and locally consistent regions but lacks global shape awareness. \textit{Geodesic Voronoi partitioning}~\cite{papadopoulou1998new} (d) uses farthest point sampling to select patch centres and assigns vertices via geodesic distance, producing uniformly distributed connected regions, though without explicitly encoding curvature or semantic structure. 
All these tokenisation policies provide a sub-optimal basis for attention mechanisms and correspondence estimation (e.g., to various irregularities or coarseness). 
These observations motivate our final design with curvature guidance, which combines curvature and spectral cues with overlapping regions to balance local detail preservation and global geometric consistency (Fig.~\ref{fig:tokenization_comparison}). 

\begin{figure}[t]
    \centering

    \begin{subfigure}[t]{0.49\linewidth}
        \centering
        \includegraphics[
        width=\linewidth,
        trim={0.6cm 0.3cm 0cm 0.6cm},
        clip
    ]{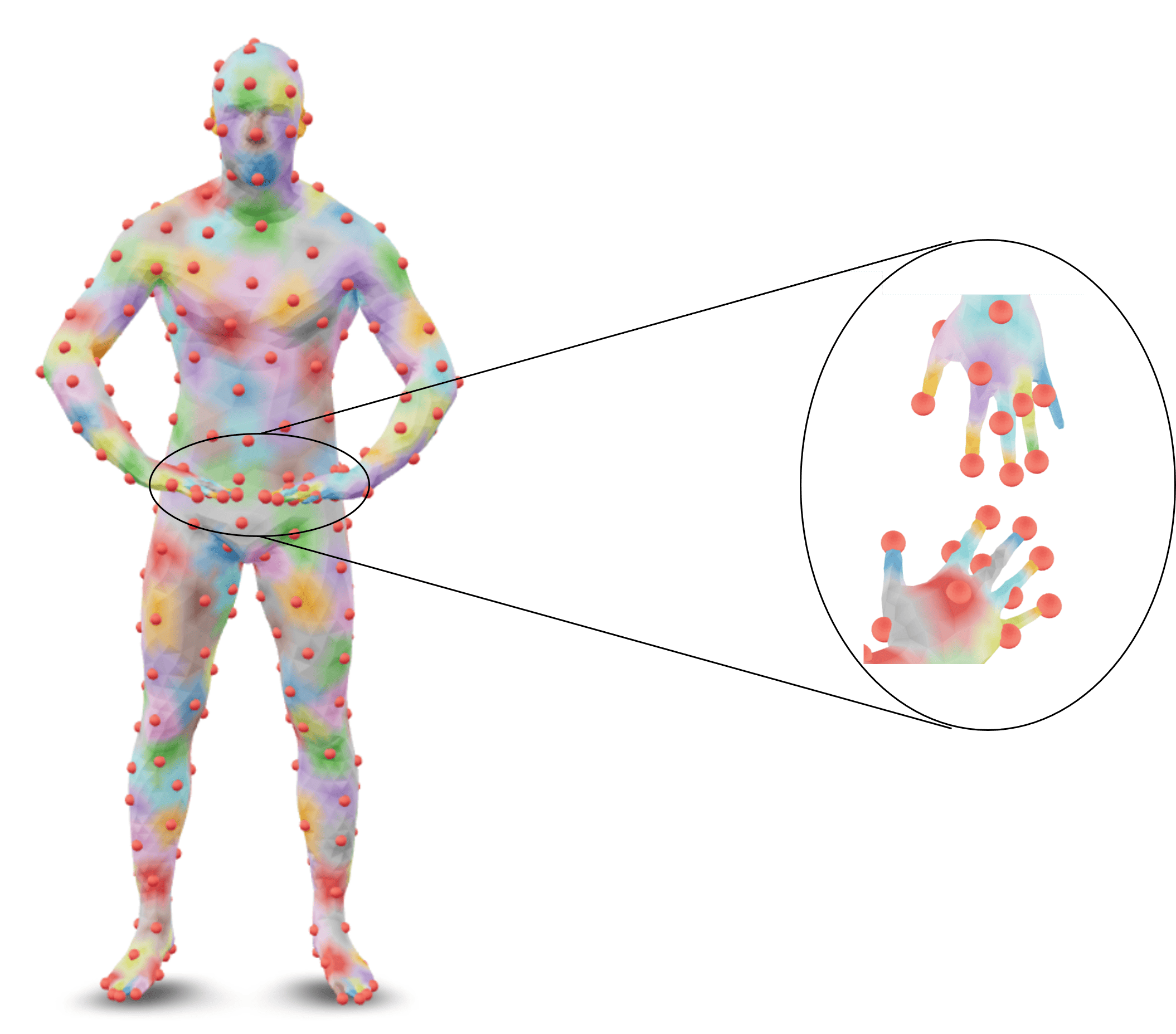}
        \caption{\textbf{Curvature-aware mesh tokenisation.}
        Tokens concentrate around high-curvature and articulated regions.}
        \label{fig:hand_zoomed}
    \end{subfigure}
    \hfill
    \begin{subfigure}[t]{0.49\linewidth}
        \centering
        \includegraphics[
            width=\linewidth,
            height=0.27\textheight,
            keepaspectratio
        ]{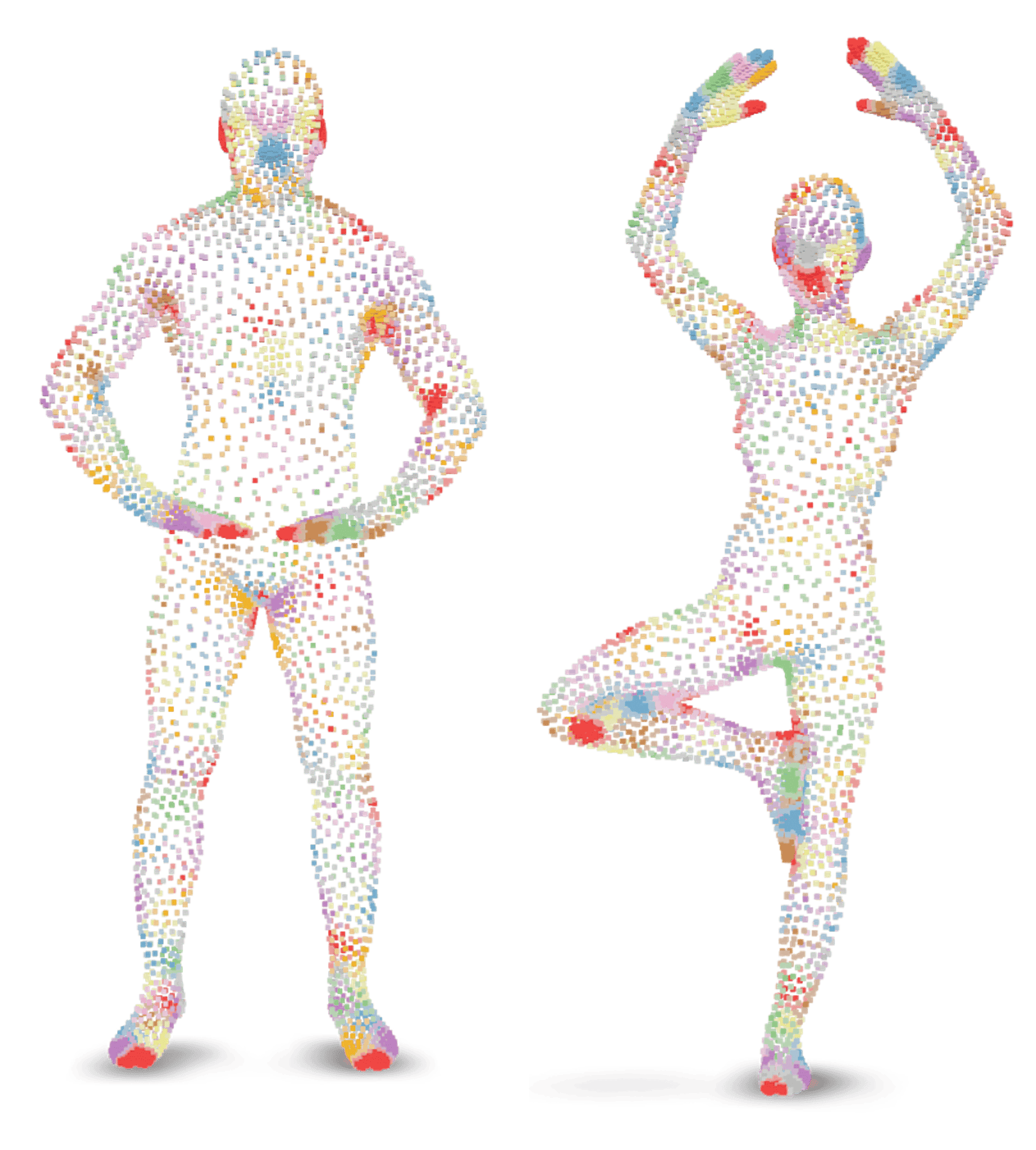}
        \caption{\textbf{Point-cloud tokenisation.}
        Token distribution over irregular surface samples without mesh connectivity.}
        \label{fig:pcd_tokens}
    \end{subfigure}

    \caption{\textbf{Token allocation on meshes and point clouds.}
    (a): The proposed curvature-aware tokenisation adaptively concentrates
    token centres on structurally informative regions, providing finer
    coverage of articulations and boundaries.
    (b): For point cloud inputs, the absence of mesh connectivity leads to
    less structured token placement and reduced concentration around
    high-curvature regions.}
    \label{fig:tokenisation_comparison}
\end{figure}

\begin{table}[!t]
\centering
\small
\caption{Comparison of different tokenisation strategies and feature representations across benchmarks. We evaluate classical tokenisation strategies and our proposed curvature-guided overlapping tokenisation with different feature encodings on the CP2P24~\cite{attaiki2021dpfm}, PSMAL~\cite{ehm2024partial}, and BeCoS~\cite{ehm2025beyond} datasets. 
All features for meshes and point clouds are learned. 
}
\label{tab:token_types}

\begin{tabular}{cccccc}
\toprule
\textbf{Method} & \textbf{Tokenisation Type} & \textbf{Feature Type} &
\textbf{CP2P24} $\uparrow$ & \textbf{PSMAL} $\uparrow$ & \textbf{BeCoS} $\uparrow$ \\
\midrule

\multirow{5}{*}{\textbf{Ours}}
& Curvature-Guided (overlapping) & Mesh
& \textbf{85.56} & \textbf{85.21} & \textbf{65.25} \\

& Spectral Laplacian~\cite{belkin2001laplacian} & Mesh
& 76.23 & 75.81 & 50.15 \\

& HKS clustering~\cite{bronstein2010scale} & Mesh
& 75.01 & 74.65 & 49.02 \\

& Face splitting~\cite{meshmae2022} & Mesh
& 75.62 & 75.21 & 51.23 \\

& Curvature-Guided (overlapping) & Point cloud
& 81.05 & 80.12 & 59.55 \\

\bottomrule
\end{tabular}
\vspace{-10pt}
\end{table}

Table~\ref{tab:token_types} evaluates the impact of different mesh tokenisation strategies while keeping the transformer architecture and learned mesh feature representation fixed. The goal of this experiment is to isolate the contribution of the proposed curvature-guided tokenisation from other framework components. The proposed curvature-guided overlapping tokenisation consistently outperforms all alternative tokenisation schemes across the CP2P24, PSMAL, and BeCoS benchmarks. 
These results indicate that the choice of tokenisation plays a critical role in the design of transformer-based shape correspondence estimation. While spectral and HKS-based approaches leverage intrinsic shape information, they partition the surface according to a fixed spectral decomposition that may not align with semantically meaningful correspondence regions. Similarly, hierarchical subdivision produces spatially regular tokens but is largely agnostic to the underlying geometric complexity of the surface. In contrast, our curvature-guided strategy allocates tokens according to local geometric variation, enabling the transformer to focus representational capacity on regions that are most informative for establishing correspondences. Furthermore, the use of overlapping tokens provides additional contextual continuity across neighbouring regions, reducing sensitivity to token boundaries and facilitating the modelling of correspondences that span multiple local surface structures. This is particularly beneficial for partial matching scenarios, where informative geometric cues may be fragmented or only partially observed. 

To better understand the local effects of the tokenisation, Fig.~\ref{fig:hand_zoomed} provides a zoomed-in visualisation of token centres on the hand region. It shows that token placement is not uniform, but instead concentrates on geometrically informative structures such as fingertips, knuckles, and boundary regions. As token density increases, these high-curvature areas are progressively refined, indicating that the method adaptively allocates representational capacity to regions of higher geometric complexity. This adaptive behaviour is particularly advantageous for correspondence estimation, as it preserves fine-grained structure in regions that are most sensitive to deformation and partial observations. Overall, these findings validate our central design choice: \textit{incorporating geometry-aware tokenisation into the transformer leads to more discriminative and transferable shape representations, yielding consistently strong performance across diverse shape matching benchmarks.} 

\paragraph{Mesh vs.~Point Cloud Representations.}
We find that our method supports point clouds with only minimal adjustments in tokenisation. 
Fig.~\ref{fig:pcd_tokens} illustrates the tokenisation of point-cloud inputs, where the lack of mesh connectivity leads to less structured coverage of high-curvature regions. To evaluate the importance of the underlying geometric representation, we compare our proposed mesh-based framework against a point-cloud variant that uses the same transformer architecture, training protocol, and correspondence objective. As shown in Table~\ref{tab:token_types} (last row), replacing the mesh representation with point clouds leads to a consistent decrease in performance across all benchmarks (still, this variant ranks second), reducing accuracy from $85.56\%$ to $81.05\%$ on CP2P24, from $85.21\%$ to $80.12\%$ on PSMAL, and from $65.25\%$ to $59.55\%$ on BeCoS. 
This result suggests that the gains achieved by our method are not solely due to the transformer backbone, but also \textit{stem from preserving the intrinsic surface structure available in meshes}. 
Meshes explicitly encode local connectivity and neighbourhood relationships, providing richer geometric context for token construction and feature learning. 
Consequently, the transformer can reason over representations that better capture both local surface characteristics and global shape structure, leading to more reliable correspondence estimation under deformation, partiality, and geometric variability. 
These findings further support our choice of meshes for generalisable shape correspondence estimation models. 

\subsection{Effect of Cross-Shape Interaction}\label{subsec:cross-atten}

We next analyse the impact of the cross-attention mechanism, which enables explicit interaction between source and target shape tokens. To isolate its effect, we remove cross-attention while keeping the curvature-guided tokenisation, mesh representation, and transformer backbone unchanged.
As shown in Table~\ref{tab:ablation_crossattn}, removing cross-attention leads to a consistent degradation in performance across all benchmarks, with accuracy dropping from $85.56\%$ to $83.94\%$ on CP2P24, from $85.21\%$ to $83.01\%$ on PSMAL, and from $65.25\%$ to $63.96\%$ on BeCoS. Self-attention alone---while effective for intra-shape reasoning---is insufficient to model all nuances of correspondence associations (with the highest accuracy) between two distinct shapes. 
\begin{table}[!h]
\centering
\caption{Effect of cross-attention in the proposed architecture.}
\label{tab:ablation_crossattn}
\begin{tabular}{lccc}
\toprule
\textbf{Variant} & \textbf{CP2P24} & \textbf{PSMAL} & \textbf{BeCoS} \\
\midrule

Full model & \textbf{85.56} & \textbf{85.21} & \textbf{65.25} \\
w/o cross-attention & 83.94 & 83.01 & 63.96 \\

\bottomrule
\end{tabular}
\end{table}
In contrast, cross-attention facilitates direct feature exchange between the source and target representations, allowing the model to establish correspondence hypotheses through joint reasoning rather than independent encoding. This becomes particularly important in partial or ambiguous settings, where local evidence alone is insufficient, and global cross-shape context is required to disambiguate matches. Overall, this experiment highlights that explicit inter-shape interaction is a key component for achieving accurate and robust shape correspondence estimation.
\begin{table}[!h]
\centering
\caption{\textbf{Effect of geodesic Gaussian bandwidth $\sigma$ on partial-to-partial shape matching.}
We vary $\sigma$ to control token overlap. Moderate overlap ($\sigma = 0.15$) yields the best performance.} 
\label{tab:sigma_ablation}

\begin{tabular}{lccc}
\toprule
\textbf{$\sigma$} & \textbf{CP2P24} $\uparrow$ & \textbf{PSMAL} $\uparrow$ & \textbf{BeCoS} $\uparrow$ \\
\midrule
0.00 (no-overlapping) & 84.05 & 83.92 & 64.04 \\
0.05 & 83.91 & 83.72 & 64.01 \\
0.10 & 84.88 & 84.63 & 64.72 \\
\textbf{0.15} & \textbf{85.56} & 85.21 & \textbf{65.25} \\
0.20 & 85.31 & \textbf{85.99} & 65.10 \\
0.25 & 84.40 & 84.10 & 64.38 \\
0.30 & 80.12 & 82.95 & 61.45 \\
\bottomrule
\end{tabular}
\end{table}
\subsection{Token Overlap Degree}\label{subsec:overlap-degree}
We study the effect of the geodesic Gaussian overlap bandwidth $\sigma$, which controls how strongly neighbouring token regions overlap in our soft tokenisation scheme. Given token centres, we compute geodesic distances from each mesh element to all centres and convert them into soft assignment weights using a Gaussian decay. $\sigma$ directly controls the spatial extent of each token. Low values produce sharply localised regions that behave similarly to hard partitions, while larger values generate more diffuse tokens with larger overlap between neighbouring regions. In practice, we observe that moderate overlap yields the most stable representations. Very small $\sigma$ values can lead to brittle assignments that are sensitive to mesh irregularities, whereas overly large values reduce spatial discriminability by over-smoothing geometric structure. This confirms that a balanced level of overlap is important for robust and geometry-consistent tokenisation, as shown in Table~\ref{tab:sigma_ablation}.

\subsection{Number of Mesh Tokens}\label{subsec:num-token}

We analyse the effect of the number of mesh tokens $g$, which controls the spatial resolution of our tokenisation. A smaller number of tokens leads to coarse surface coverage, while increasing $g$ allows the model to capture finer geometric details. 
\begin{wrapfigure}{r}{0.5\textwidth}
  \centering
  \includegraphics[
      width=\linewidth,
      trim={0 0.7cm 0 2.5cm},
      clip
  ]{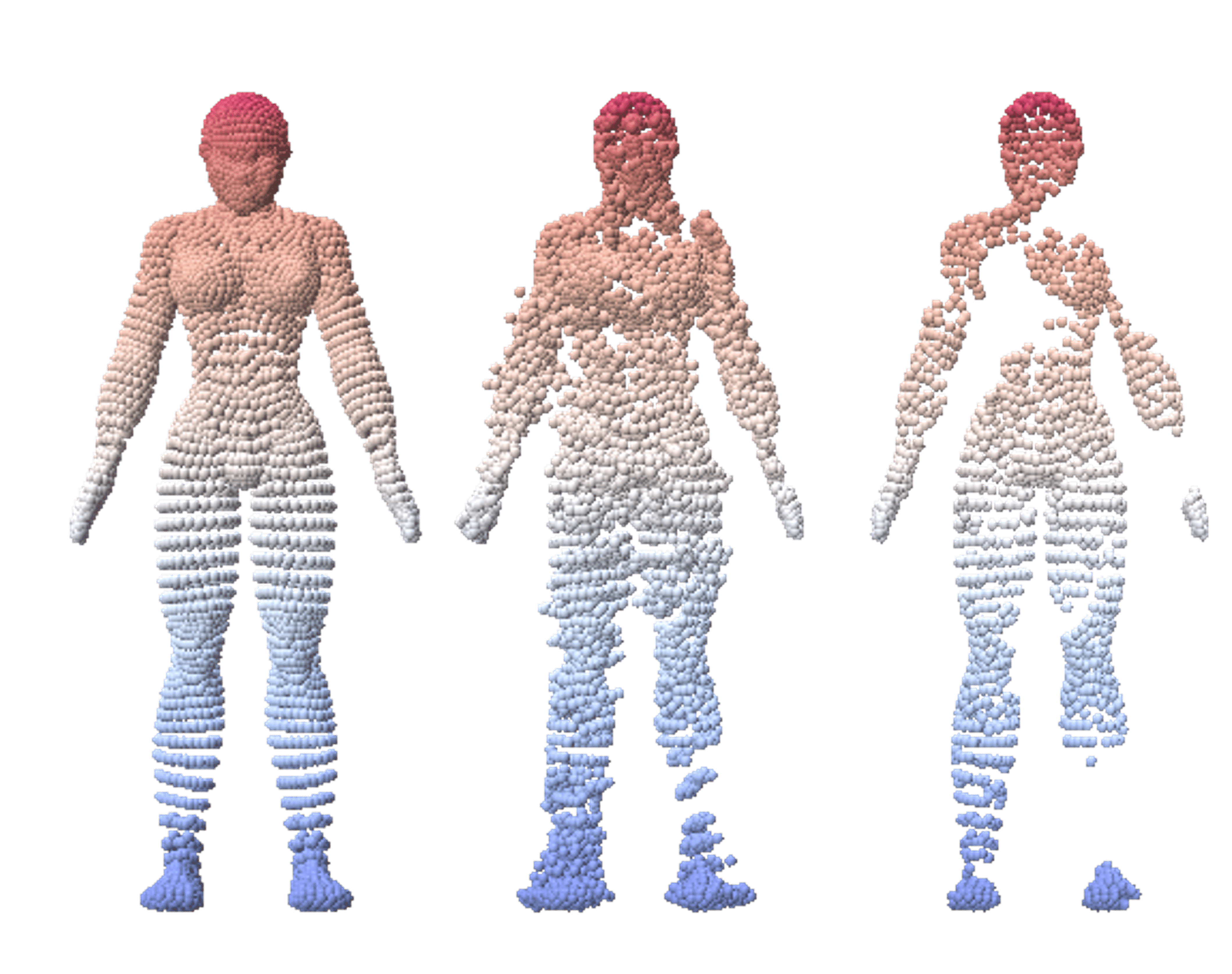}
  \caption{We show ground-truth shapes (left), reconstructions produced by our model (centre), and masked inputs (right). Our model recovers missing geometric structures from heavily occluded inputs while preserving fine-grained details and global shape consistency.}
  \label{fig:masking}
  \vspace{-40pt}
\end{wrapfigure}

We evaluate our method with $g \in \{32, 64, 128, 256\}$ while keeping all other components fixed. As shown in Table~\ref{tab:token_count_ablation}, increasing $g$ consistently improves performance by enabling better coverage of local geometric structures. However, gains begin to saturate at higher resolutions, indicating a trade-off between representational granularity and redundancy.

\begin{table}[t]
\centering
\caption{\textbf{Effect of number of mesh tokens $g$ on partial-shape matching.}
Increasing $g$ improves coverage of fine geometric details but shows diminishing returns at higher resolutions.}
\label{tab:token_count_ablation}
\vspace{3pt}
\begin{tabular}{c|ccc}
\toprule
\textbf{$g$} & \textbf{CP2P24} $\uparrow$ & \textbf{PSMAL} $\uparrow$ & \textbf{BeCoS} $\uparrow$ \\
\midrule
32  & 75.20 & 70.75 & 55.91 \\
64  & 78.15 & 72.60 & 61.45 \\
128 & 83.72 & 81.32 & 62.14 \\
256 & \textbf{85.56} & \textbf{85.21} & \textbf{65.25} \\
\bottomrule
\end{tabular}
\end{table}

\subsection{Impact of Token Masking Ratio}\label{ssec:masking_ratio}

\setlength{\intextsep}{3pt}

Our empirical findings show that masked auto-encoder pre-training (Sec.~\ref{sec:mae}) provides a substantial performance boost and improves the robustness and generalisation of the learned representations. 
To study the effect of the masking ratio in our masked auto-encoding pre-training strategy, we vary the proportion of mesh tokens masked during training, which forces the model to reconstruct missing geometric information from partial shapes. We evaluate masking ratios of 0.5, 0.7, and 0.8 while keeping all other components fixed. As shown in Table~\ref{tab:mask_ablation}, moderate masking provides the best balance between learning robust global structure and preserving sufficient context for reconstruction; see Fig.~\ref{fig:masking}. Excessive masking reduces available geometric context, making learning overly difficult.

\begin{table}[!h]
\centering
\caption{\textbf{Effect of token masking ratio in masked auto-encoding pre-training.}
Moderate masking ($50\%$) yields the best performance by balancing reconstruction difficulty and geometric context.}
\label{tab:mask_ablation}
\vspace{3pt}
\begin{tabular}{c|ccc}
\toprule
\textbf{Mask Ratio} & \textbf{CP2P24} $\uparrow$ & \textbf{PSMAL} $\uparrow$ & \textbf{BeCoS} $\uparrow$ \\
\midrule
0.10 & 81.24 & 83.01 & 62.15 \\
0.30 & 82.75 & 84.18 & 64.21 \\
0.50 & \textbf{85.56} & \textbf{85.21} & \textbf{65.25} \\
0.70 & 83.15 & 80.22 & 59.71 \\
0.80 & 80.21 & 82.02 & 60.04 \\
\bottomrule
\end{tabular}
\end{table}

\newpage
\begin{figure}[h]
    \centering
    \vspace{80pt}
    \begin{subfigure}[t]{0.95\linewidth}
        \centering
        \includegraphics[width=\linewidth]{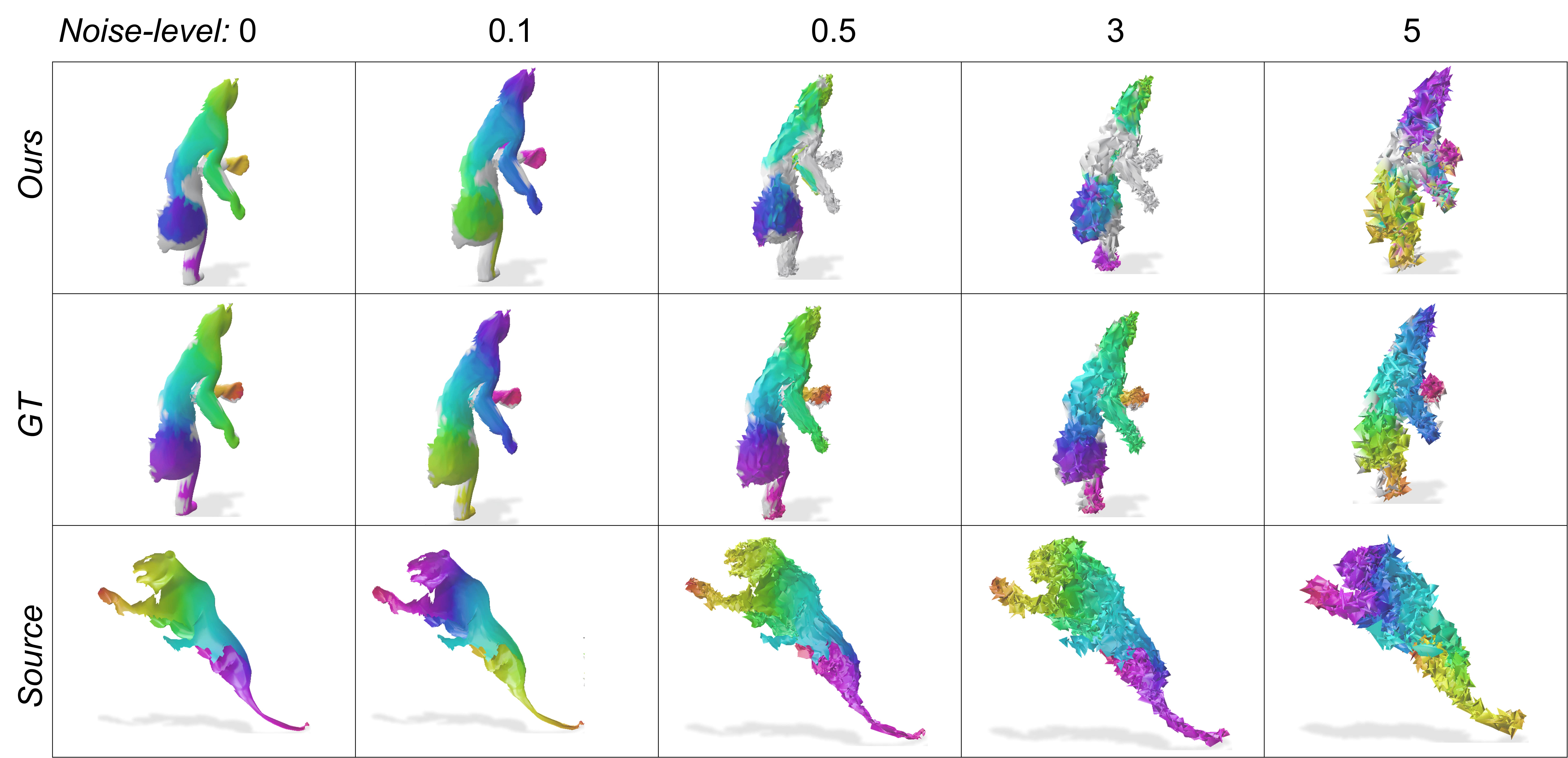}
    \end{subfigure}

    \vspace{2.3em}

    \begin{subfigure}[t]{0.95\linewidth}
        \centering
        \includegraphics[width=\linewidth]{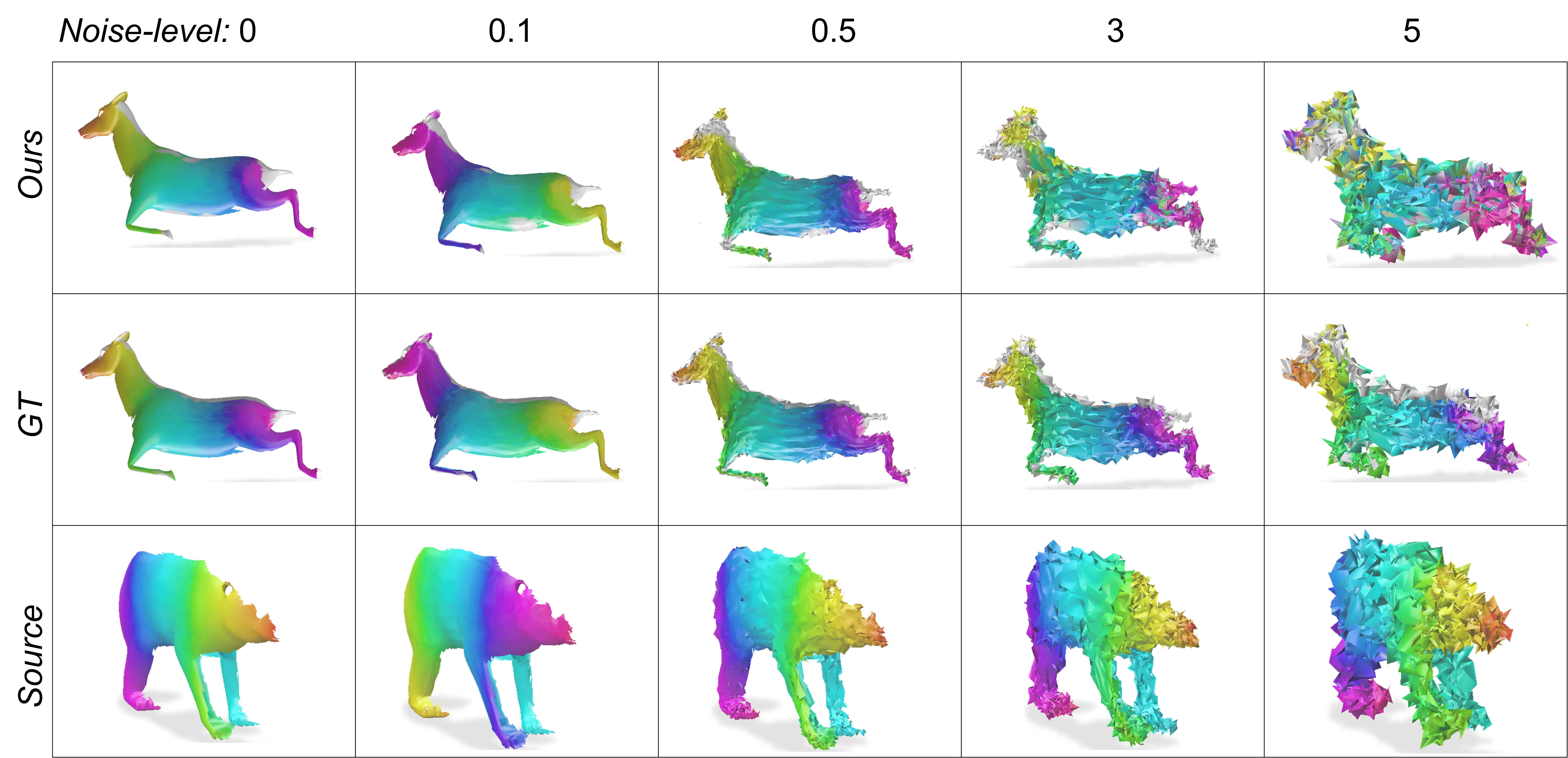}
    \end{subfigure}

    \vspace{0.7em}

    \caption{\textbf{Robustness to geometric noise on the BeCoS \cite{ehm2025beyond} dataset.}
    Shown are results on two different partial-to-partial shape pairs under increasing vertex noise levels ($0.1$, $0.5$, $3.0$ and $5.0$). Despite strong geometric perturbations (both rightmost columns), correspondences estimated by TokenMatch degrade gradually; the method preserves structurally consistent correspondences and remains robust even under severe noise.} 
    \label{fig:noise_robustness}
\end{figure}

\newpage
\begin{figure}[h]
  \centering
  \includegraphics[width=\linewidth]{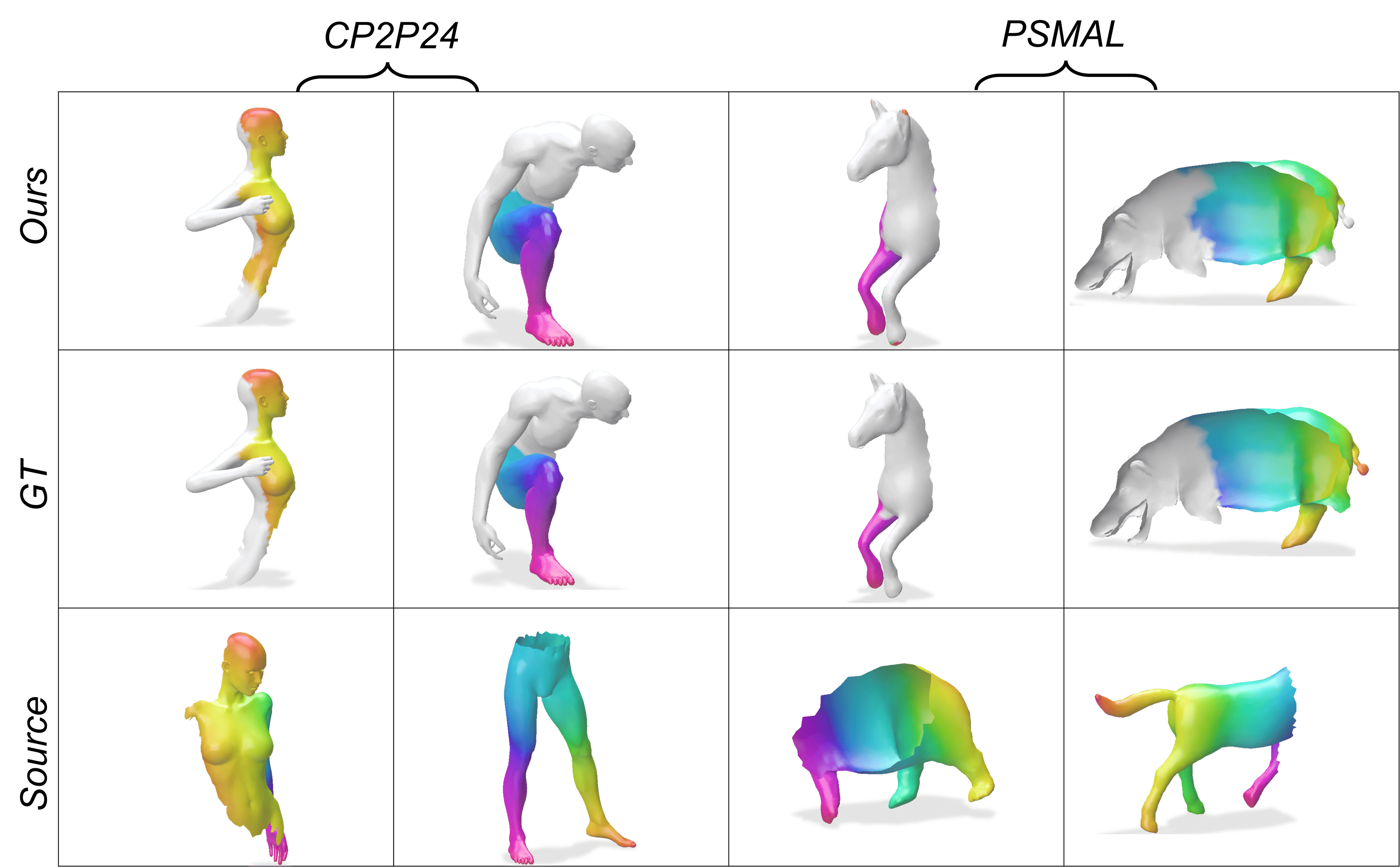}
  \caption{\textbf{Additional qualitative results for challenging scenarios on the CP2P24 and PSMAL datasets} 
  under varying levels of partiality and deformation. Matches are colour-coded \textit{columnwise}.}
  \label{fig:qualadd1}
\end{figure}
\vspace{20pt}
\begin{figure}[h]
  \centering
  \includegraphics[width=\linewidth]{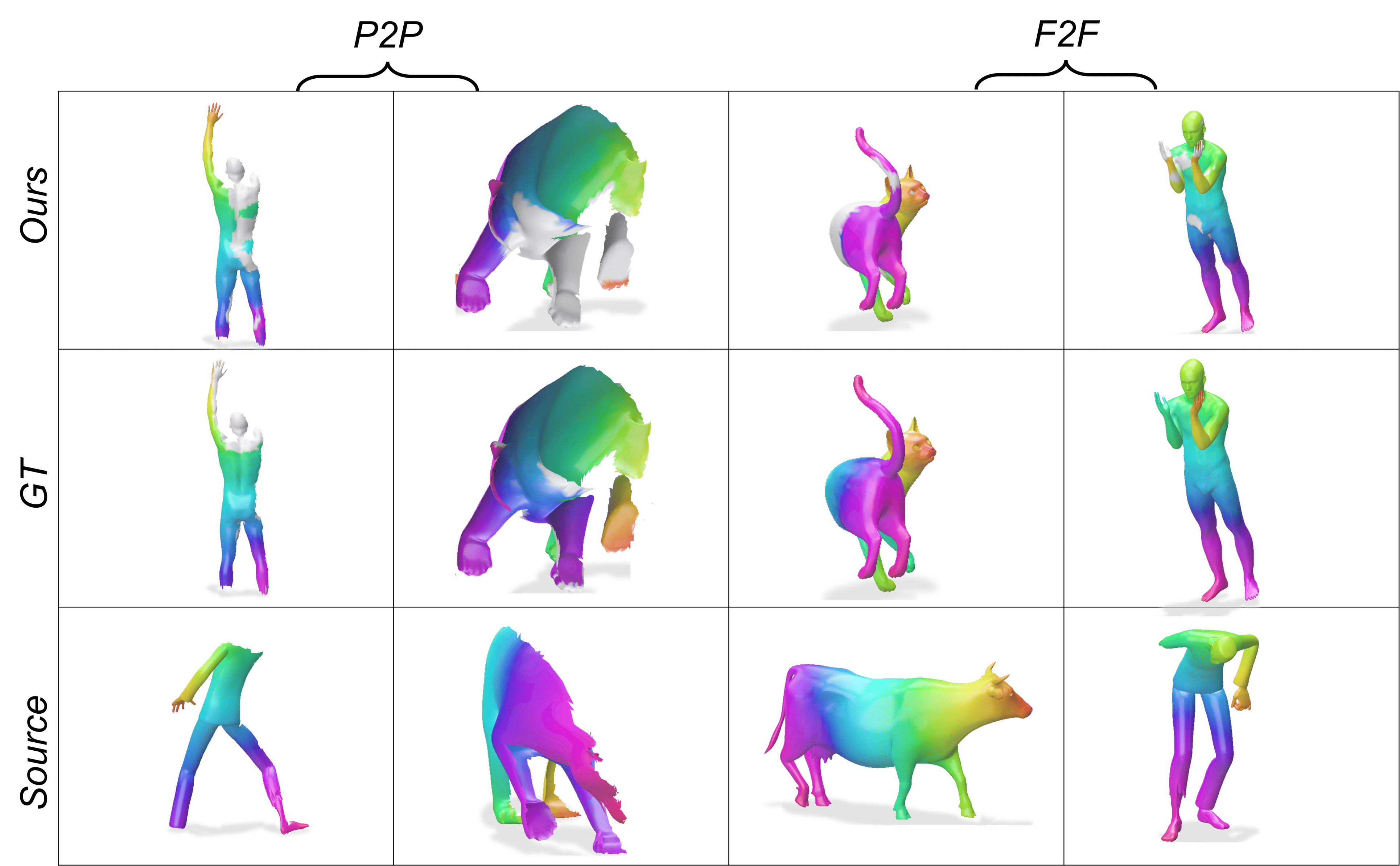}
  \caption{\textbf{Additional qualitative results on BeCoS \cite{ehm2025beyond}:} The model is exclusively trained on partial shapes (partial-to-partial setting denoted by ``P2P'') and generalises well to full shapes (full-to-full setting denoted by ``F2F''). Matches are colour-coded \textit{columnwise}. ``GT'' stands for ground truth.} 
  \label{fig:qualadd2}
\end{figure}

\newpage
\begin{figure}[th!]
    \centering
    \includegraphics[width=1.0\linewidth]{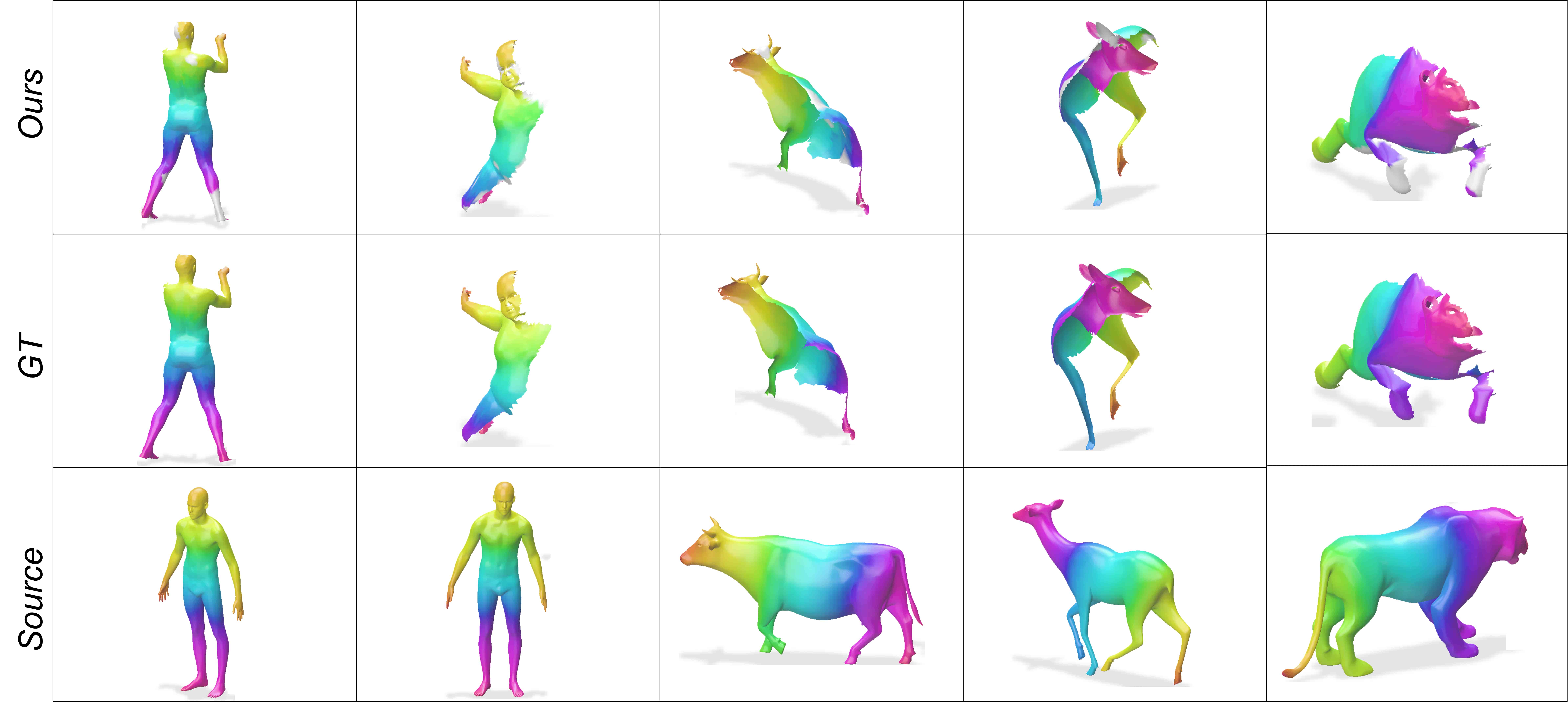}
    \caption{\textbf{Generalisation from partial-to-partial training to partial-to-full evaluation on the BeCoS \cite{ehm2025beyond} dataset.}
    Our model produces consistent and semantically meaningful correspondences across complete shapes, demonstrating high robustness and transferability of the learned geometry-aware representations.} 
    \label{fig:becos_generalization}
\end{figure}

\subsection{Robustness to Noise}\label{ssec:noise_robustness}
We evaluate the robustness of our method under synthetic geometric perturbations applied at test time. Importantly, \textit{the model is trained exclusively on clean meshes, and noise is introduced only during evaluation to simulate real-world scanning artefacts and reconstruction imperfections.} 

{\setlength{\columnsep}{6pt}
\begin{wraptable}{r}{0.30\textwidth}
\centering
\footnotesize
\setlength{\tabcolsep}{2.5pt}
\renewcommand{\arraystretch}{0.9}
\vspace{-2pt}
\caption{Method robustness under Gaussian noise $\sigma_n$.} 
\label{tab:noise_robustness}

\begin{tabular}{lcc}
\toprule
\textbf{$\sigma_n$} &
\makecell{\textbf{Mean IoU}\\\textbf{($\times100$) $\uparrow$}} &
\makecell{\textbf{Drop}\\\textbf{($\Delta$)}} \\
\midrule
$0.0$ (No noise) & 65.25 & \textemdash \\
$0.1$            & 64.73 & $-0.52$  \\
$0.5$            & 62.38 & $-2.87$  \\
$3.0$            & 48.22 & $-17.03$ \\
$5.0$            & 42.45 & $-22.80$ \\
\bottomrule
\end{tabular}
\vspace{-15pt}
\end{wraptable} 
We generate noisy meshes by perturbing vertex positions with Gaussian noise. Specifically, for each mesh, we compute the bounding box diagonal to normalise scale, and then add zero-mean Gaussian noise with standard deviation $\sigma_n$ proportional to the chosen noise level. This ensures that the perturbation is scale-invariant across different shapes. We consider multiple noise levels by varying $\sigma_n$ relative to the mesh scale. 
Table~\ref{tab:noise_robustness} shows that performance decreases by only 0.52 points (the third column reports the performance drop $\Delta$) under mild noise and 2.87 points under moderate noise, indicating robustness to geometric perturbations without noise-specific training. 
Fig.~\ref{fig:noise_robustness} presents qualitative results under the same increasing geometric noise levels (0.1, 0.5, 3, and 5) on the partial-to-partial BeCoS \cite{ehm2025beyond} dataset. While severe noise causes expected degradation, TokenMatch remains robust under mild and moderate perturbations, highlighting an additional dimension of generalisability beyond the training data manifold. 

\section{Further Qualitative Results}
\label{sec:additional_qualitative}

We provide additional qualitative results on the CP2P24 and PSMAL datasets to further illustrate the performance of our method under varying levels of partiality and non-isometric deformation in Figs.~\ref{fig:qualadd1} and~\ref{fig:qualadd2}. 
The visualisations demonstrate that our TokenMatch produces accurate and geometrically consistent correspondences even in challenging partial-to-partial settings, where shape overlap is limited and local structure is highly incomplete.
We further present results on BeCoS, which contains more realistic and unconstrained scenarios with significant noise, missing regions, and strong deformations. Notably, as shown in Fig.~\ref{fig:becos_generalization}, our model effectively handles full shapes and settings with varying degrees of partiality. 
The predicted correspondences remain stable across large geometric variations, preserving semantic alignment even under substantial shape deformation. 
Overall, these qualitative results highlight the robustness of our approach and its ability to learn transferable geometric representations that generalise beyond the training distribution.

\end{document}